\documentclass[11pt]{article}

\usepackage[final]{acl}
\usepackage{times}
\usepackage{latexsym}
\usepackage{xcolor}
\usepackage[T1]{fontenc}
\usepackage[utf8]{inputenc}

\usepackage{microtype}

\usepackage{inconsolata}
\usepackage{graphicx}
\usepackage{multirow}
\usepackage{enumitem}
\usepackage{pgfplots}
\pgfplotsset{compat=1.18}
\usepackage{rotating}
\usepackage{subcaption}

\pgfplotsset{
    significance/.style={
        nodes near coords,
        nodes near coords align=center,
        every node near coord/.append style={
            font=\scriptsize\bfseries,
            yshift=6pt,
            rotate=90      % <-- THIS makes it vertical
        }
    }
}

\title{BALANCER: \textbf{B}alanced \textbf{A}gentic \textbf{L}earning via \textbf{A}daptive dy\textbf{N}amic \textbf{C}ontext \textbf{E}xt\textbf{R}action for Hard Word Sense Disambiguation}
\author{
    Deshan Sumanathilaka, Nicholas Micallef, Julian Hough \\
    Department of Computer Science, Swansea University, Wales, UK \\
    { \{t.g.d.sumanathilaka, nicholas.micallef, julian.hough\}@swansea.ac.uk}
}

\begin{document}
\maketitle
\begin{abstract}
%The advent of Transformer-based architectures has substantially advanced natural language understanding, particularly in modeling contextual meaning and resolving ambiguity. 
Large Language Models (LLMs), trained on trillions of tokens, demonstrate strong capabilities in capturing common lexical semantics. 
% However, these models continue to struggle with less frequent homonymous word senses, which are often underrepresented in training data and consequently receive insufficient attention during inference.
However, these models still struggle with less frequent homonyms, which are underrepresented in training data and receive limited attention during inference. As a result, misinterpretations of such senses can propagate substantial downstream errors, especially in domain-specific %, technical, or high-stakes 
  contexts.
% (e.g., "charge" denoting an accusation rather than a monetary fee in legal documents).
We propose an Agentic LLM-driven pipeline that dynamically retrieves and integrates external knowledge to enhance sense interpretation. The proposed framework promotes balanced consideration of all possible word senses during the reasoning process, enabling more accurate identification of the intended meaning in context. We benchmark our approach against two-step fine-tuned, low-parameter LLM baselines and show that enforcing balanced sense representation significantly improves Word Sense Disambiguation (WSD) performance. Our method achieves state-of-the-art performance on rare and hard senses (HardEN \& 42D) benchmarks. In addition, we introduce HardMWSD, a curated subset of XLWSD designed to stress-test edge cases across languages. Experimental results confirm that the proposed Agentic approach generalizes effectively in multilingual settings, establishing a robust and scalable solution for fine-grained sense disambiguation.

\end{abstract}

\section{Introduction}

Generative AI has revolutionized day-to-day tasks by transforming complex processes into natural, text-driven systems. Tools such as Alexa, ChatGPT, and Gemini have become integral to everyday workflows, facilitating conversational interactions that help streamline human activities~\cite{aslitdinova2025artificial}. However, accurate language understanding is crucial, as a single misunderstanding of a word can lead to significant misinformation and incorrect conclusions~\cite{chen_poswsd_2022}.

% Changed the order
Working with polysemous words remains a major challenge for current NLP systems. Highly polysemous words are difficult to disambiguate because their senses are unevenly distributed across contexts, with a small number of meanings dominating general usage while many others occur only in narrow domains. For example, in WordNet~\cite{wordnet}, the word \emph{``shot''} has several senses, including the act of firing a projectile, a medical injection, and attempting to score in sports, as well as domain-specific meanings.
% such as a missile or spacecraft launch in aerospace contexts, an explosive charge used in blasting in mining, the shot-put implement in athletics, and a cinematic shot in film production.

For instance, in aerospace, the term \emph{``shot''} may refer to a spacecraft or missile launch attempt, which differs from its more common meaning of firing a projectile. Consider the sentence ``The final shot was delayed due to instability in the upper-stage ignition sequence.'' Without sufficient contextual detail, ``shot'' could be misinterpreted as a weapon discharge rather than a launch attempt. This example shows how the domain-specific use of highly polysemous words can lead to ambiguity, underscoring the importance of contextual understanding for accurately identifying the intended sense. To identify the exact meaning, word sense disambiguation (WSD) algorithms play a key role. Although WSD is a component of NLP pipelines, recent work on transformer-based architectures has found that while these models perform well on general and frequent senses, they struggle with rare and infrequent senses~\cite{meconi-etal-2025-large}.

Thus, in this research, we investigate how to enhance the disambiguation of rare senses using an Agentic LLM-based approach that leverages real-world information to improve sense interpretation.~\citet{sumanathilaka2025prompt} found that imbalanced sense distributions and biases toward commonly used meanings can cause models to over-predict frequent senses, resulting in limited attention to rare cases, particularly in multilingual WSD.~\citet{zhang-etal-2025-rodeval} identified that incorrect sense selections do not cause ``forgetting'' correct options, but are due to incomplete acquisition of all senses for polysemous words and inherent biases. 

To address these issues, we propose a dynamic sense information extraction mechanism based on Agentic modelling, designed to streamline the disambiguation process. Unlike prior supervised or fine-tuned WSD systems that rely on static representations, our novelty lies in dynamically enforcing balanced sense evidence at inference time, using an Agentic retrieval verification loop. The goal is not to outperform large supervised systems, but to improve robustness on rare and hard senses, especially when model capacity or labelled data is limited. The key contributions of this research are:
% \begin{enumerate}[leftmargin=*,nosep]
%     \item We propose and evaluate an Agentic LLM-based pipeline that enhances the interpretation of rare senses by promoting balanced attention across all senses during disambiguation.
%     \item We establish new state-of-the-art results on the HardEN and 42D benchmarks, which are recognized as challenging datasets for English WSD.
%     \item We introduce HardMWSD, a curated challenging subset of XLWSD, designed to evaluate generalizability in non-English languages.
% \end{enumerate}

\begin{enumerate}[leftmargin=*,nosep]
    \item We introduce BALANCER, an Agentic framework for \textit{balanced evidence construction} that dynamically retrieves, verifies, and equalizes contextual evidence across competing word senses. Unlike conventional retrieval-augmented WSD, the framework explicitly mitigates lexical frequency bias by enforcing balanced sense representations during inference, improving the disambiguation of rare/domain-specific meanings.

    \item We provide an empirical analysis demonstrating that balancing external sense evidence substantially improves long-tail WSD beyond conventional fine-tuning. Specifically, we show that while curriculum-based fine-tuning strengthens representations of frequent senses, balanced evidence construction is critical for resolving rare and difficult lexical ambiguities.

    \item We achieve new state-of-the-art performance on the challenging HardEN and 42D benchmarks while introducing HardMWSD, a multilingual benchmark derived from XLWSD for evaluating difficult and low-frequency sense disambiguation across multiple languages, showing that our framework generalizes beyond English.
\end{enumerate}

%The remainder of this paper is structured as follows: the Background and Related Work section reviews the evolution of WSD and relevant prior studies; the Methodology section describes the approach adopted in this work; this is followed by the presentation of results and observations, and finally, the Conclusions and Directions for Future Research section.
In the following sections, we review related work on WSD, outline our methodology, present our results, and conclude with future research.

\section{Background and Related Work}

\subsection{Evolution of WSD}

Accurate analysis of word meaning is essential for many NLP tasks. Machine translation, question answering, and text summarization all rely on effective WSD, as incorrect sense selection can lead to erroneous or misleading outputs~\cite{abeysiriwardana2024survey}. Beyond system performance, such errors can cause significant real-world consequences in high-impact societal domains, including healthcare, public safety, and misinformation detection~\cite{satpute2024rule}. In these settings, misinterpreting a key term may lead to flawed clinical guidance or amplify misleading information~\cite{awotunde2025word}. 

As a result, research has focused on improving semantic understanding in NLP systems. Prior work has explored rule-based~\cite{bowerman1978acquisition,palmer2006evaluation} approaches as a fundamental WSD approach, which were later extended to knowledge-based techniques using structured lexical resources such as WordNet, ontologies, and semantic networks~\cite{chaplot_knowledge-based_2018,agirre2009personalizing,basile2014enhanced}. With the creation of large datasets such as SemCor~\cite{miller1994using}, supervised techniques were prominent, with transformer-based methods achieving SOTA performance in WSD, including HardWSD~\cite{song_improved_2021,huang_glossbert_2020}.

\subsection{Large Language Models for WSD}

Although recent work has substantially improved LLM performance on common vocabulary, evidence shows that models continue to struggle with rare and domain-specific ambiguous terms~\cite{cahyawijaya2024thank}. One contributing factor is model scale.~\citet{yae2024leveraging} demonstrate that WSD accuracy improves with increasing model size, though larger models entail substantially higher computational costs. In contrast, parameter-efficient models may lack the rich contextual representations compared to large-scale LLMs. However, fine-tuning these models for downstream WSD tasks can significantly improve their performance over their base versions~\cite{basile2025exploring}. Beyond fine-tuning, several studies explore the reformulation of WSD to better leverage LLM capabilities. \citet{sainz_what_2023} cast WSD as a textual entailment task, prompting models to evaluate the compatibility between candidate domain labels and sentences containing ambiguous words. This zero-shot approach outperforms random baselines and, in some cases, matches or exceeds supervised WSD systems. Related work investigates prompt engineering and in-context learning with GPT-3.5-Turbo and GPT-4-Turbo~\cite{sumanathilaka2024assessing}, while subsequent benchmarking identifies Deepseek R1 and GPT-4-mini as effective for WSD~\cite{sumanathilaka_can_2024}. 

Additionally,~\citet{meconi-etal-2025-large} extend evaluation to the hardEN and 42D datasets using various LLMs, reporting their limitations in handling hard cases. However, they also show that, when the target word is explicitly specified, commercial models achieve high accuracy in generating examples.~\citet{sumanathilaka2025prompt} report that an uneven few shots can cause performance degradation due to the close attention to the frequent senses.
Together, these trends show that WSD remains an active research area, evolving from a standard pipeline component to a targeted challenge even in the era of large language models.

Thus, we adopt an Agentic LLM framework that explicitly guides the model’s attention towards the intended lexical item and dynamically retrieves sense-relevant contextual evidence. Rather than relying on implicit sense inference, our approach decomposes the WSD process into structured Agentic steps that enforce sense awareness and balance focus across both frequent and infrequent senses. 

% These findings further motivated us to use the Agentic LLM framework to enhance the sense interpretation.

\subsection{Agentic LLM and its advancements}

Recent advances in LLMs, particularly in reasoning capabilities, have opened new avenues for NLP tasks. Unlike conventional predictive models, Agentic LLMs aim to simulate aspects of human cognitive processes by reasoning, acting, and interacting with their environment. These agents iteratively acquire new information, reason over it, and make decisions accordingly~\cite{plaat2025Agentic}. Agentic LLM frameworks have recently become popular across several NLP tasks, demonstrating improved performance for planning~\cite{webb2025brain} and verification~\cite{lee2025Agentic}.

In the context of WSD, recent work~\cite{zhang-etal-2025-madawsd} introduces a multi-agent debating framework composed of specialized roles, including a Debater, Moderator, Consensus-Seeker, and Judge, to simulate more accurate sense resolution. Their approach is evaluated on adversarial Chinese and English datasets~\cite{ballout2024fool}, demonstrating notable effectiveness. Also, many other sensitive domains where language understanding plays a major role, such as legal~\cite{petros-etal-2025-pakton}, medical~\cite{shang2026dynamicare}, and communication~\cite{li2026reinforcement}, have shown promising improvements through multi-agent frameworks. Despite the growing adoption of Agentic LLMs across diverse domains, their application to WSD remains underexplored. In this study, we leverage recent advances in Agentic frameworks to enhance sense interpretation by incorporating knowledge exploration and verification agents, thereby streamlining the disambiguation process and specifically targeting hard cases.

\section{Methodology}
This section describes the experimental methodology used in this study, including the datasets and LLMs employed for English and multilingual WSD evaluation. We then present our two-step fine-tuning approach, followed by the Agentic LLM pipeline designed to address data imbalance and improve disambiguation of rare and domain-specific senses.

\subsection{Dataset selection}

We used various English and multilingual WSD datasets. For English, we used the FEWS dataset~\cite{blevins2021fews}, a few-shot dataset derived from Wiktionary containing 121K labelled examples for training, including 5K records each for zero-shot and few-shot testing. We also used SemCor~\cite{miller1994using}, the largest manually sense-annotated corpus aligned with WordNet, comprising 226K sense annotations. In addition, we used the ALLamended dataset~\cite{maru_nibbling_2022}, which combines Senseval-2 and 3, and SemEval 2013 and 2015. For multilingual evaluation, we used XLWSD~\cite{pasini2021xl}, a cross-lingual benchmark, limiting our experiments to French, German, Italian, and Spanish to ensure a controlled and representative evaluation across Romance and Germanic languages. To assess model robustness, we used hardEN~\cite{maru_nibbling_2022}, which comprises 476 challenging test cases that existing state-of-the-art (SOTA) systems have not solved, and 42D~\cite{maru_nibbling_2022}, a challenge dataset containing 370 instances with rare and domain-specific senses.

We curated a HardMWSD (Model-Challenging Stress Set), introducing a multilingual, hard-WSD stress set of 1,205 records that are difficult for current SOTA low-parameter LLMs for MWSD to disambiguate, drawn from the XLWSD dataset. HardMWSD is not a balanced benchmark but a targeted stress test for rare and difficult cases, with 141 (de), 360 (es), 225 (fr) and 479 (it) records that were carefully sense-mapped using BabelNet~\cite{navigli2010babelnet}. We have provided the dataset\footnote{\href{https://github.com/Sumanathilaka/HardMWSD_Dataset/}{HardMWSD dataset}} and Appendix~\ref{sec:appendixc} with more details about the HardMWSD. The required candidate keys are retrieved from WordNet 3.0 and BabelNet.

\subsection{LLMs used in this research}
We used various LLMs from each family during each phase. The two-step fine-tuning was performed on low-parameter LLMs (\textless 4B) including Qwen-3-4B ~\cite{yang2025qwen3}, Gemma3-4B \cite{team2025gemma}, SmolLM3-3B ~\cite{bakouch2025smollm3}, Llama-3.2-3B and for the Agentic LLM-based pipeline, we used non-fine-tuned open-source low-parameter LLMs and commercial LLMs including GPT-4o~\cite{gpt4o}, Gemini 2.5 Flash Lite~\cite{gemini-flash-lite}, and Deepseek V3~\cite{liu2024deepseek,deepseek-v3}. 

\subsection{Two-step fine-tuning approach}
\label{2phaseFT}

Our main objective was to fine-tune low-parameter models for the WSD task using a two-step curriculum learning (incrementally fine-tuning LLMs from simple to complex tasks) to teach word-sense identification~\cite{bengio2009curriculum}. \\

\noindent\textbf{Step 1: Teaching semantics}%\\
    
In this step, our objective was to train the models to infer and disambiguate the correct sense of an ambiguous word directly from context, without relying on explicit sense ID detection or classification. For English, we fine-tuned the base models using the FEWS dataset. Multilingual corpora translated from SemCor and WNG were used~\cite{pasini2021xl}. Fine-tuning was performed using Low-Rank Adaptation (LoRA)~\cite{hu2022lora} to enable parameter-efficient adaptation to preserve the core representations of the pretrained models. The prompts and further discussion on fine-tuning are provided in Appendix \ref{AppendixAprompt}.\\

\noindent\textbf{Step 2: Word sense extraction} %\\ 
    
The second step aimed to explicitly guide the models to leverage the surrounding contextual information and Part-of-Speech (POS) tags to accurately identify the intended meaning of an ambiguous word and select the correct sense ID from a predefined sense inventory. Fine-tuning was guided by a CoT-based prompt that required selecting the most suitable meaning. We further fine-tuned the models obtained from step one using an instruction-based formulation tailored for sense selection. 

\noindent \textbf{Study Setup.} We adopted a two-stage curriculum-based fine-tuning setup. In the first stage, models were trained for 2 epochs with a learning rate of 2e-4, using AdamW (weight decay 0.01) and a linear scheduler with 50 warmup steps. A batch size of 4 with gradient accumulation over 8 steps was used to ensure memory efficiency, and mixed-precision (FP16) training was enabled. In the second stage, models were fine-tuned for 2 epochs with a lower learning rate of 1e-5 to mitigate catastrophic forgetting and preserve previously acquired semantic knowledge, using AdamW (weight decay: 0.01) and a cosine scheduler to ensure stable updates. Hyperparameters in both stages were selected empirically to balance training stability, convergence, and sense disambiguation performance.

\subsection{Agentic LLM for enhancing sense representation}

Although the proposed two-step fine-tuning approach shows strong performance on common senses and a subset of rare cases, it still struggles with extremely low-frequency and corner-case senses. Our analysis indicates that this limitation is largely related to imbalanced training data, where infrequent senses receive insufficient representation and attention during model optimization. This imbalance is detrimental when disambiguating domain-specific and rare senses, which require more explicit and diversified contextual evidence.

\begin{figure}
    \centering
    \includegraphics[scale=0.28]{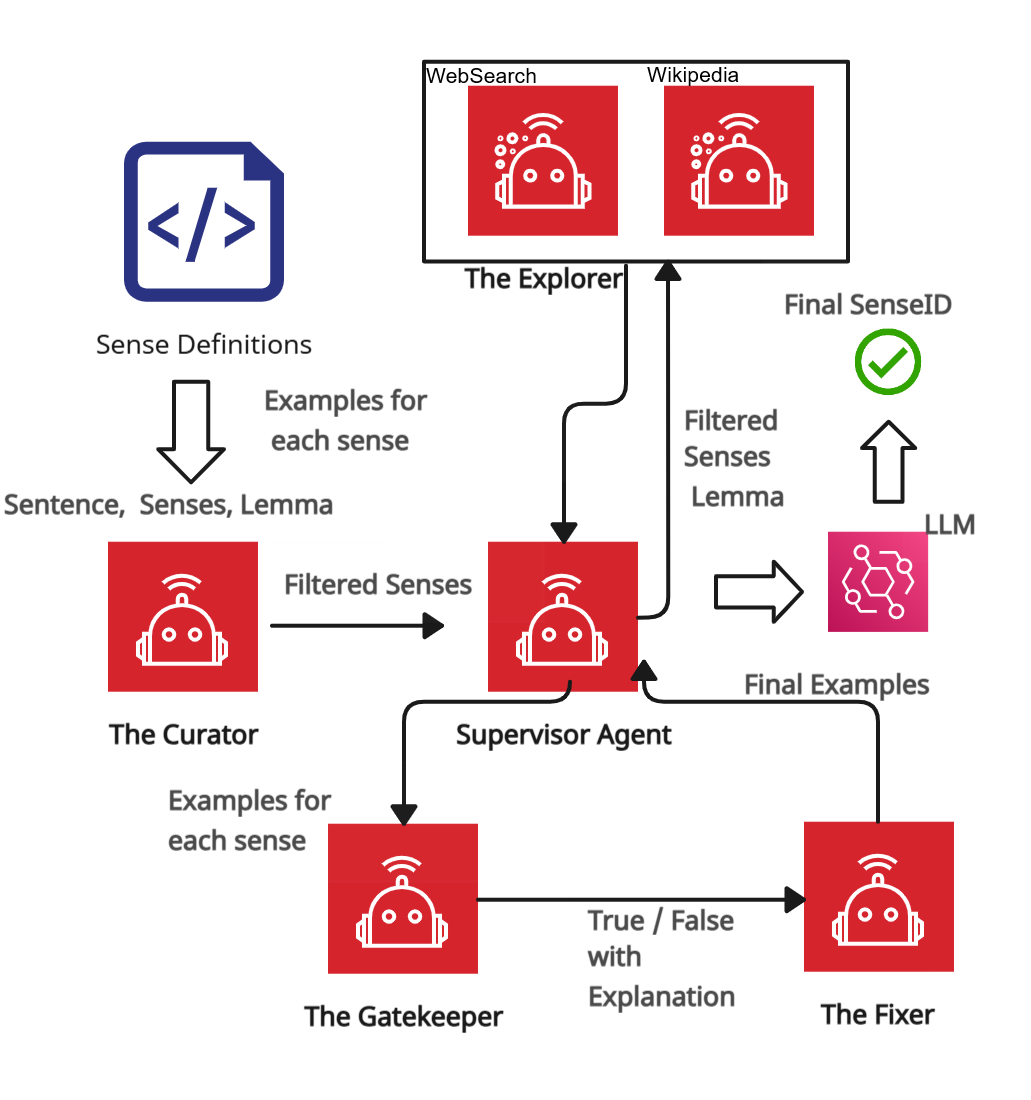}
    %\caption{Flow of the proposed Pipeline with Agentic LLMs}
    \caption{Proposed Agentic LLM Pipeline used for the BALANCER framework.}
    \label{fig:AgenticPipeline}
\end{figure}

Prior work~\cite{sumanathilaka2025prompt} has similarly shown that imbalanced few-shot examples can lead to incorrect sense predictions in multilingual settings, as models tend to overemphasize high-frequency senses at the expense of less frequent alternatives. Thus, we propose an Agentic LLM-based pipeline to dynamically extract and integrate external and contextual knowledge, enabling a more balanced representation of all candidate senses during the reasoning process. The proposed pipeline has five specialized agents (Figure \ref{fig:AgenticPipeline}), each responsible for a distinct yet complementary function within the WSD process. We developed the architecture using a human-in-the-loop approach, where iterative human assessment and refinement informed the selection of the most effective pipeline configuration~\cite{zhu2026hler}. A detailed discussion on the human involvement is presented in Appendix~\ref{sec:appendixarch}.

\textbf{Sense Filtering Agent (The Curator):} This agent serves as the first-stage pruning mechanism for the sense space. Given an ambiguous word, its surrounding sentence context, and a predefined list of candidate sense IDs, this agent performs semantic compatibility analysis to eliminate irrelevant senses while retaining potentially valid or borderline cases. Importantly, the agent adopts a high-recall filtering strategy where senses are excluded only when they are semantically contradictory to the contextual usage, whereas partially relevant or uncertain senses are preserved. To support robust filtering for rare and domain-specific senses, the agent is equipped with access to external knowledge via web search, enabling it to consult dictionaries, encyclopedic resources, and lexical databases.  
    
\textbf{Web Search Agent (The Explorer):} This agent acts as an information expansion component and is responsible for enriching each retained sense with evidence. For every candidate sense ID produced by the Sense Filtering Agent, the explorer retrieves two diverse contextual examples for English and three examples for Multilingual setups that explicitly instantiate the intended meaning, as in prior work~\cite{sumanathilaka2025prompt}. These examples are sourced from authoritative web-based resources, such as Wikipedia, Merriam-Webster, Britannica, Dictionary.com and the Oxford Dictionary, to maximize semantic coverage. For domain-specific cases, websites such as BBC News and The New York Times (see the full list of resources used in the Appendix \ref{sec:appendixc}) are used. By enforcing a fixed number of examples per sense, this agent plays a key role in sense equalization, ensuring that rare senses are represented with the same amount of evidence as frequent ones. This mitigates the tendency of LLMs to over-prioritize dominant senses simply due to their prevalence in pretraining data.

\textbf{Verification Agent (The Gatekeeper):} The Verification Agent serves as a quality control and constraint enforcement mechanism. Its main role is to verify the structural and semantic completeness of the evidence produced by the Web Search Agent. Specifically, it checks that each sense ID is associated with exactly the required number of valid, coherent examples. Rather than modifying content directly, the Gatekeeper produces a binary validation signal accompanied by feedback when constraints are violated. This explicit verification step prevents the introduction of incomplete or biased evidence.

\textbf{Correction Agent (The Fixer):} When the Verification Agent signals an issue, the Correction Agent is activated to repair and regenerate the sense evidence. Guided by the verification feedback, this agent retrieves missing or incorrect examples and reconstructs the full set of sense representations in a consistent and standardized format. By overwriting previous results, the Correction Agent ensures that downstream components always operate on a clean, validated representation of the sense space.  The Gatekeeper and the Fixer operate in a bounded feedback loop with at most two iterations. This limit was selected based on preliminary experiments, which showed that the first correction pass achieved an overall success rate of 0.82, while a second pass increased it to nearly 0.99. As the second iteration produced a substantial improvement and additional iterations offered limited practical benefit, a maximum of two iterations was adopted as an effective trade-off between reliability and computational efficiency.

\textbf{Supervisor Agent (The Moderator):}
The Supervisor Agent acts as a decision-level controller that determines which agent to invoke and whether to continue the iterative refinement loop, as illustrated in Figure~\ref{fig:AgenticPipeline}. From an implementation perspective, the Supervisor is designed as a sequential agent that includes an Explorer Agent and a LoopAgent (using Google ADK), which orchestrates the Gatekeeper and Fixer Agents. This centralized control mechanism ensures coherence across agent outputs, prevents over-reliance on any single information source, and maintains the data quality required for robust disambiguation.
    
The proposed Agentic pipeline transforms WSD from a single-pass prediction problem into a multi-stage reasoning process that explicitly addresses data imbalance, rare-sense under representation, and domain specificity. All agents in the pipeline were implemented using Gemini-based models to support the intermediate reasoning and data exploration stages. However, the final disambiguation step in the BALANCER pipeline was performed using a base model without fine-tuning, ensuring that the model had no access to web-based tools.

\noindent \textbf{Study Setup.} For Agentic pipeline generation, we used the Google Agent Development Kit (ADK)~\cite{google-adk}, with Gemini 2.5 Flash Lite serving as the core reasoning model for each agent. The framework included external tools, such as Google Search, to support information retrieval and decision-making. Agents were assigned a temperature of 0 and a fixed seed to ensure reproducibility. For the inference process, we used the GLOSSGPT prompt template, based on few-shot Chain-of-Thought (CoT)~\cite{sumanathilaka_glossgpt_2025}.  
Temperatures followed model recommendations, with 0 used for commercial LLMs~\cite{temperature_2024_deshan}. We used Together AI~\cite{together-ai} for the Deepseek-V3 model, and Azure~\cite{microsoft-foundry} for GPT-4o models.

\section{Findings and Discussion}

\subsection{Phase 1: Two-step fine-tuning}
 
In Phase 1, we evaluated model performance using the proposed two-step fine-tuning strategy (Section~\ref{2phaseFT}) to assess how effectively models acquire WSD knowledge during sequential training with SemCor and FEWS.

Table~\ref{tab:baseline_vs_our} compares the two-step fine-tuned models against their base model using zero-shot prompting. The results show that while the proposed approach achieves better performance on ALLamended benchmarks and XLWSD, its performance on hardEN and 42D remains lower than that of the baseline models. This highlights a key limitation: models trained on general-purpose corpora tend to overfit frequent or dominant senses, reducing sensitivity to rare senses and harming performance on datasets with difficult or low-frequency cases.

In contrast, on the ALLamended dataset, the proposed approach consistently outperforms the baseline low-parameter LLMs. This improvement indicates that two-step fine-tuning substantially enhances sense representations for common and moderately frequent senses, reinforcing the effectiveness of continued fine-tuning in strengthening semantic knowledge. However, the degradation observed on harder benchmarks underscores an important insight: fine-tuning alone, when driven by imbalanced sense distributions, is insufficient to ensure robust disambiguation across all senses. These findings reinforce the need for mechanisms that balance focus across senses, particularly those underrepresented during training.
This served as the main motivation for the next phase of the study.

\DeclareRobustCommand{\legendbox}[1]{
\tikz{\protect\draw[fill=#1,line width=1pt]  (0cm,0cm) rectangle ++(0.25cm,0.25cm)}
}

\begin{table}[t]
\centering
\small
\caption{Model performance (F1 Score) on hardEN, 42D, and ALLamended(ALL) datasets. FT: Fine-tuned.}
\label{tab:baseline_vs_our}

\begin{tabular}{l@{\hspace{3mm}}c@{\hspace{3mm}}c@{\hspace{3mm}}c@{\hspace{3mm}}c}
\hline
\textbf{Model} & \textbf{hardEN} & \textbf{42D} & \textbf{ALL} & \textbf{XLWSD}  \\
\hline
\multicolumn{4}{c}{\textbf{Baseline : Without fine-tuning}} \\
\hline
Qwen3-4B        & 34.2 & 47.1 & 66.1 & 73.87\\
Gemma3-4b       & 34.5 & 62.9 & 72.8 & 72.18\\
SmolLM3-3B       & 29.4 & 43.2 & 46.5 & 63.11 \\
\hline
\multicolumn{4}{c}{\textbf{2 phase fine-tuning}} \\
\hline
FT Qwen3-4B    & 10.36 & 23.39 & 76.16 & 88.23
\\
FT Gemma3-4B  & 12.60 & 29.89 & 76.15 & 85.02
\\
FT SmolLM3-3B  & 17.62 & 35.56 & 76.63 & 79.19
\\
\hline
\end{tabular}

\end{table}

\begin{table*}[t]
\centering
\small
\caption{Performance comparison of Two-step fine-tuning using F1 score across languages with XLWSD dataset}
\label{tab:multilingual_results}
\resizebox{\textwidth}{!}{
\begin{tabular}{lcccccccc}
\hline
\multirow{2}{*}{\textbf{Language}} 
& \multicolumn{5}{c}{\textbf{Recent Works}} 
& \multicolumn{3}{c}{\textbf{Two-Step Fine Tuning}} \\
\cline{2-9}
& \textbf{SyntagRank} 
& \textbf{EWISER} 
& \textbf{XLMR} 
& \textbf{ConSeC} 
& \textbf{SANDWICH} 
& \textbf{Gemma} 
& \textbf{Qwen} 
& \textbf{SmolLM} \\
\hline
French  & 70.0 & 80.9 & 83.9 & 84.4 & \textbf{92.8} & 85.21 & 89.31 & 79.76 \\
German  & 76.0 & 80.9 & 83.1 & 84.2 & \textbf{93.2} & 85.98 & 90.89 & 81.76 \\
Italian & 69.6 & 74.6 & 77.6 & 79.3 & 86.6 & 85.50 & \textbf{88.25} & 79.76 \\
Spanish & 68.6 & 71.9 & 75.9 & 77.4 & 84.0 & 84.13 & \textbf{86.48} & 77.70 \\
\hline
\end{tabular}
}
\end{table*}

\begin{figure*}[t]
\centering
% hardEN Dataset
\begin{subfigure}[b]{0.32\textwidth}
\centering
\begin{tikzpicture}
\begin{axis}[
    ybar,
    bar width=0.10cm,
    width=\textwidth,
    height=5cm,
    enlarge x limits=0.15,
    % legend style={at={(0.5,-0.25)},
    %     anchor=north,legend columns=1, font=\tiny},
    ylabel={F1 Score},
    ylabel style={font=\small},
    title style={font=\small\bfseries},
    symbolic x coords={Qwen, Gemma, SmolLM3, Gemini, Deepseek, GPT-4o},
    xtick=data,
    xlabel style={font=\small},
    x tick label style={rotate=45, anchor=east, font=\tiny},
    ymin=0,
    ymax=100,
    ytick={0,15,30,45,60,75,90},
    grid=major,
    grid style={dashed,gray!30},
]

% Baseline data 
\addplot[fill=blue!60] coordinates {
    (Qwen, 10.36)
    (Gemma, 12.60)
    (SmolLM3, 17.62)
    (Gemini, 48.11)
    (Deepseek, 39.50)
    (GPT-4o, 45.60)
};

% WSF
\addplot[fill=red!60, significance, point meta=explicit symbolic ] coordinates {
    (Qwen, 57.49) [***] 
    (Gemma, 49.61) [***] 
    (SmolLM3, 52.86) [***] 
    (Gemini, 62.54) [***] 
    (Deepseek, 67.04) [**] 
    (GPT-4o, 65.92) [**] 
};

\addplot[fill=green!60, significance, point meta=explicit symbolic] coordinates {
    (Qwen, 54.43) [***] 
    (Gemma, 39.73) [***] 
    (SmolLM3, 45.20) [***] 
    (Gemini, 59.59) [***] 
    (Deepseek, 65.54)[**] 
    (GPT-4o, 66.10) [**] 
};

\end{axis}
\end{tikzpicture}
\caption{hardEN Dataset}
\end{subfigure}
\hfill
% 42D Dataset
\begin{subfigure}[b]{0.32\textwidth}
\centering
\begin{tikzpicture}
\begin{axis}[
    ybar,
    bar width=0.10cm,
    width=\textwidth,
    height=5cm,
    enlarge x limits=0.15,
    % legend style={at={(0.5,-0.25)},
    %     anchor=north,legend columns=1, font=\tiny},
    ylabel={F1 Score},
    ylabel style={font=\small},
    xlabel style={font=\small},
    title style={font=\small\bfseries},
    symbolic x coords={Qwen, Gemma, SmolLM3, Gemini, Deepseek, GPT-4o},
    xtick=data,
    x tick label style={rotate=45, anchor=east, font=\tiny},
    ymin=0,
    ymax=100,
    ytick={0,15,30,45,60,75,90},
    grid=major,
    grid style={dashed,gray!30},
]

% Baseline data (42D Baseline)
\addplot[fill=blue!60] coordinates {
    (Qwen, 23.39)
    (Gemma, 29.89)
    (SmolLM3, 35.56)
    (Gemini, 70.54)
    (Deepseek, 74.60)
    (GPT-4o, 76.81)
};

% WSF data (42D WSF)
\addplot[fill=red!60, significance, point meta=explicit symbolic] coordinates {
    (Qwen, 75.84) [***] 
    (Gemma, 69.26) [***] 
    (SmolLM3, 69.26) [***] 
    (Gemini, 83.60) [**]
    (Deepseek, 83.28) [**] 
    (GPT-4o, 85.96)[*] 
};

% WOSF data (42D WOSF)
\addplot[fill=green!60, significance, point meta=explicit symbolic] coordinates {
    (Qwen, 73.42) [***] 
    (Gemma, 59.85) [***] 
    (SmolLM3, 62.20) [***] 
    (Gemini, 81.60) [**]
    (Deepseek, 83.10) [**] 
    (GPT-4o, 85.98)[*] 
};

\end{axis}
\end{tikzpicture}
\caption{42D Dataset}
\end{subfigure}
\hfill
% hardMWSD Dataset
\begin{subfigure}[b]{0.32\textwidth}
\centering
\begin{tikzpicture}
\begin{axis}[
    ybar,
    bar width=0.10cm,
    width=\textwidth,
    height=5cm,
    enlarge x limits=0.15,
    % legend style={at={(0.5,-0.25)},
    %     anchor=north,legend columns=1, font=\tiny},
    ylabel={F1 Score},
    ylabel style={font=\small},
    title style={font=\small\bfseries},
    symbolic x coords={Qwen, Gemma, SmolLM3, Gemini, Deepseek, GPT-4o},
    xtick=data,    
    xlabel style={font=\small},
    x tick label style={rotate=45, anchor=east, font=\tiny},
    ymin=0,
    ymax=100,
    ytick={0,15,30,45,60,75,90},
    grid=major,
    grid style={dashed,gray!30},
]

% Baseline data (hardMWSD Baseline)
\addplot[fill=blue!60] coordinates {
    (Qwen, 0)
    (Gemma, 0)
    (SmolLM3, 0)
    (Gemini, 41.25)
    (Deepseek, 46.80)
    (GPT-4o, 54.38)
};

% WSF data (hardMWSD WSF)
\addplot[fill=red!60, significance, point meta=explicit symbolic] coordinates {
    (Qwen, 51.69) [***] 
    (Gemma, 50.03) [***] 
    (SmolLM3, 54.38) [***] 
    (Gemini, 53.85) [*]
    (Deepseek, 54.03) [*]
    (GPT-4o, 56.12)[*] 
};

% WOSF data (hardMWSD WOSF)
\addplot[fill=green!60, significance, point meta=explicit symbolic] coordinates {
    (Qwen, 44.52) [***] 
    (Gemma, 44.32) [***] 
    (SmolLM3, 48.14) [***] 
    (Gemini, 56.29) [*]
    (Deepseek, 58.07) [*]
    (GPT-4o, 62.44)[**] 
};

\end{axis}
\end{tikzpicture}
\caption{HardMWSD Dataset}
\end{subfigure}

%\begin{tikzpicture}
%\small
%\begin{axis}[
%    hide axis,
%    xmin=0, xmax=1,
%    ymin=0, ymax=1,
%    legend style={
%        draw=none,
%        at={(0.5,0.5)},
%        anchor=center,
%        legend columns=3,
%        column sep=1cm,
%        /tikz/every even column/.append style={column sep=1cm}
%    },
%    legend entries={Baseline, WSF, WOSF}
%]
%\addlegendimage{area legend, fill=blue!60}
%\addlegendimage{area legend, fill=red!60}
%\addlegendimage{area legend, fill=green!60}
%\end{axis}
%\end{tikzpicture}

%\caption{F1 Score comparison across different models on three datasets: hardEN, 42D, and hardMWSD. Each subfigure shows performance for Baseline, WSF (With Sense Filtering), and WOSF (Without Sense Filtering) configurations. Statistical Significance score vs. baseline: * $p \leq 0.05$ ,\**  $p \leq 0.01$ ,\;
%*** $p \leq 0.001$. Refer to Table \ref{tab:sense_filtering} for the stats.}
\caption{F1 score comparison of models on three datasets (hardEN, 42D, HardMWSD). The above subfigures report \legendbox{blue!60} Baseline (Zero-shot), \legendbox{red!60} WSF (With Sense Filtering), and  \legendbox{green!60} WOSF (Without Sense Filtering). Statistical significance vs. Baseline: * $p \leq 0.05$, ** $p \leq 0.01$, *** $p \leq 0.001$. See Table~\ref{tab:sense_filtering} for details.}
\label{fig:f1_comparison_all_datasets}
\end{figure*}
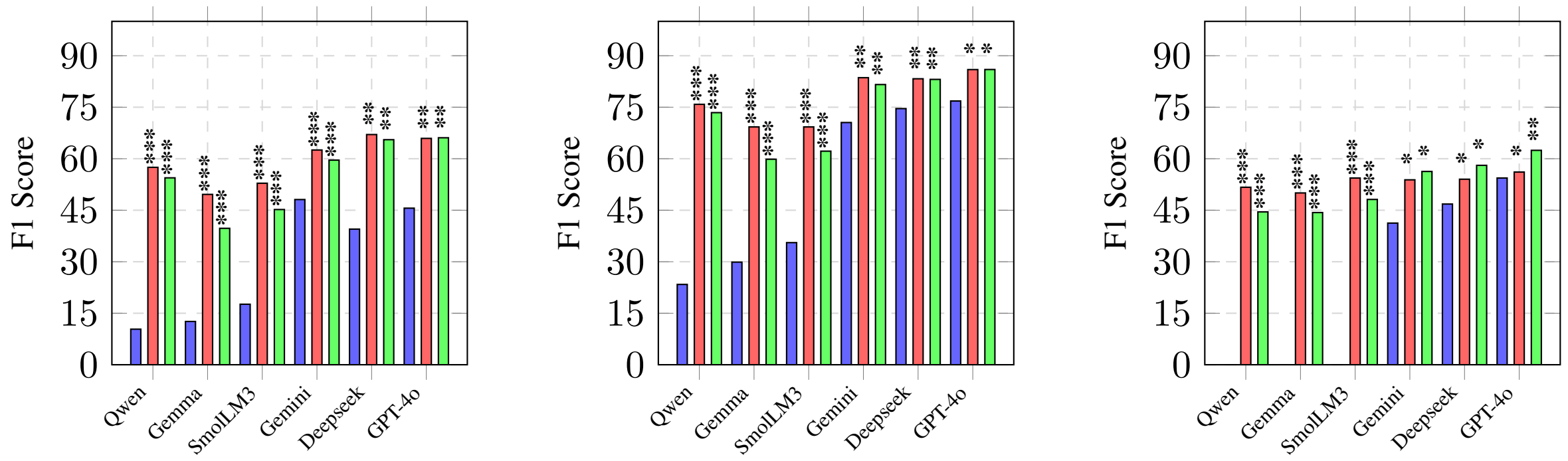

\begin{table*}[t]
\centering
\small
\caption{F1 score comparison: ARES~\cite{scarlini-et-al-2020-ares}, BEM~\cite{blevins_moving_2020}, ESC~\cite{barba_esc_2021}, EWS~\cite{bevilacqua2020breaking}, GEN~\cite{bevilacqua2020generationary}, SYN~\cite{scozzafava-etal-2020-personalized}), CSC~\cite{barba_consec_2021}, GLOSSGPT(GGPT)~\cite{sumanathilaka_glossgpt_2025}, SandWiCH~\cite{guzman2025sandwich} and EAD~\cite{sumanathilaka-etal-2026-exploration}. The best-performing approach with agentic WSD is in Italics.}

\label{tab:baseline_comparison}
\begin{tabular}{lcccccccccccc}
\hline
\textbf{Dataset} & \textbf{ARES} & \textbf{BEM} & \textbf{ESC} & \textbf{EWS} & \textbf{GEN} & \textbf{SYN} & \textbf{CSC} & \textbf{GGPT}  & \textbf{SandWiCH} & \textbf{EAD}& \textbf{\textit{BALANCER}} \\
\hline
42D     & 41.8 & 53.2 & 58.9 & 43.9 & 50.2 & 32.8 & 56.6 & 70.27 & 77.1 & 78.48 & \textbf{85.98}\\
hardEN  & 0.0  & 0.0  & 0.0  & 0.0  & 0.0   & 0.0  & 7.35 & 34.66 & 53.4 & 54.19 & \textbf{67.04}\\
\hline
\end{tabular}
\end{table*}

\begin{table*}[t]
\centering
\small
\caption{F1 score of ablation studies on the different component analysis with 42D dataset}
\label{tab:ablation on components}

\begin{tabular}{lcccc}
\hline
\textbf{Model} & \textbf{Baseline(Zero Shot)} & \textbf{GlossGPT with KB} & \textbf{Web Access agent only} & \textbf{BALANCER}\\
\hline

GPT-4o        & 76.81  & 70.27 & 74.32 & 85.98 \\
Gemini-2.5    & 70.54 & 61.35 & 72.43 & 81.60 \\

\hline
\end{tabular}
\end{table*}

To assess the generalizability of the proposed approach on other languages, we evaluated the two-step fine-tuning approach on the XLWSD multilingual benchmark. Table~\ref{tab:multilingual_results} reports performance (micro F1) compared against established SOTA systems: SyntagRank~\cite{scozzafava-etal-2020-personalized}, EWISER~\cite{bevilacqua2020breaking}, XLM-R~\cite{pasini2021xl}, ConSeC~\cite{barba_consec_2021}, and SANDWiCH~\cite{guzman2025sandwich} using same protocols as~\citet{pasini2021xl}.

We found that for Italian and Spanish, the two-step fine-tuned Qwen model achieves state-of-the-art performance, surpassing or matching all competing approaches. For French and German, the model achieved performance comparable to SANDWiCH, the current SOTA system. These findings suggest that sequential fine-tuning transfers and enriches sense knowledge even in multilingual %and comparatively low-resource 
settings. Overall, the results reinforce the importance of continuous fine-tuning to enhance semantic representations beyond English-centric training.

Despite these gains, our analysis reveals a persistent set of corner cases that remain unresolved across all fine-tuned models, regardless of language or dataset. These instances typically involve subtle semantic distinctions, rare senses, or highly context-dependent interpretations that cannot be reliably resolved through static fine-tuning alone. To analyze such failures, we curated the HardMWSD challenging set, composed of instances that none of fine-tuned models (SOTA performance on XLWSD) could correctly disambiguate, motivating our decision to base the filtering on their outputs.

These observations motivate the use of an Agentic LLM-based approach. Unlike static fine-tuned models, an Agentic framework enables iterative reasoning, dynamic context exploration, and sense comparison, allowing the model to balance evidence across competing interpretations. This capability is important for corner cases, where surface-level contextual cues are insufficient, and a multi-step reasoning process is required.

\subsection{Phase 2: Agentic-based approach}

This phase targets hard and low-frequency sense disambiguation. Thus, we evaluate only challenging English and multilingual benchmarks designed to stress-test WSD systems.
%Since the primary goal of this phase is to address hard and low-frequency sense disambiguation, we evaluate our approach exclusively on challenging benchmarks designed to stress-test WSD systems in both English and multilingual settings.
% This approach was primarily evaluated on challenging benchmarks for both English and multilingual WSD. 
Figure~\ref{fig:f1_comparison_all_datasets} reports results on the hardEN, 42D, and HardMWSD datasets, and compares our approach against strong commercial LLMs, including Deepseek-V3 and GPT-4o (using the prompt proposed by~\citet{meconi-etal-2025-large}). The comparison shows that the performance gains from our approach are considerable %pronounced 
even with low-parameter LLMs as the final decision maker. Table~\ref{tab:baseline_comparison} compares our best-performing Agentic models against supervised WSD models.

An analysis of intermediate outputs revealed that, in fewer than 8\% of instances in English and 10\% in the multilingual setup, the gold sense was incorrectly removed during the initial sense-filtering stage. To assess the impact of this filtering step, we conducted ablation experiments evaluating model performance both with and without sense filtering. This analysis provides insight into the trade-off between candidate space reduction and sense coverage. On the English WSD benchmarks, our proposed approach achieves SOTA performance on both the hardEN and 42D datasets, outperforming all previously reported systems as noted in Table~\ref{tab:baseline_comparison}. Notably, low-parameter models augmented with the proposed Agentic pipeline outperform the Deepseek-V3 and GPT-4o zero-shot baselines on hardEN, while achieving comparable performance on 42D. Furthermore, when Deepseek-V3, Gemini-2.5-Flash, and GPT-4o are used as final decision-making agents, we observed substantial F1-score improvements across all datasets with sense filtering: +27.54, +14.43, and +20.32 on hardEN; +8.68, +13.06, and +9.17 on 42D. On HardMWSD without sense filtering, it shows improvements of +11.27, +15.04, and +8.06, respectively, over the above LLMs. Importantly, with commercial LLMs, Balancer achieves better performance in the HardMwsd without the curator agent, a pattern not observed in 42D and HardEN. 

We evaluated statistical significance using McNemar’s test~\cite{mcnemar1947note} on paired instance-level predictions, comparing each model against its corresponding baseline. Figure~\ref{fig:f1_comparison_all_datasets} shows that all low-parameter models with the Agentic approach had statistically significant improvements against their baseline. Gemini 2.5-Flash-Lite showed strong significance for hardEN, with weaker significant gains for 42D and HardMWSD. GPT-4o shows significant improvements over hardEN and 42D, with mixed yet significant results on HardMWSD. Deepseek-V3 achieves statistically significant improvements on all evaluated datasets. These results demonstrate that the proposed Agentic pipeline significantly outperforms the baselines.\\

\textbf{\emph{Comparative analysis of the sense filtering stage}} indicates that low-parameter models benefit more from candidate sense pruning, whereas commercial LLMs exhibit similar performance with or without filtering. This suggests that the proposed Agentic filtering mechanism effectively compensates for limited model capacity. To further examine the generalizability of the proposed Agentic pipeline to other languages, we evaluated the approach on German, Spanish, French, and Italian using the HardMWSD benchmark, which comprises difficult instances for existing WSD systems. As reported in Table~\ref{tab:sense_filtering}, the overall results on HardMWSD indicate that employing SmolLM3-3B as the final decision-making agent yields superior performance among low-parameter models, outperforming alternative compact LLMs. This approach has significantly enhanced the performance of low-parameter LLMs, demonstrating its effectiveness. As for commercial LLMs, GPT-4o performed best, achieving an F1 score above 65 without sense filtering. As we found for English, low-parameter models benefit substantially from the initial sense-filtering stage, whereas results for commercial LLMs remain in a similar range regardless of filtering.

Table~\ref{tab:sense_filtering_lan} presents a language-specific breakdown of the Agentic evaluation results. The analysis demonstrates that low-parameter models gain measurable improvements from sense filtering, whereas the impact on larger models is comparatively limited. McNemar’s test supports this finding for larger models, showing that predictions without sense filtering are not significantly different from those with filtering (p > 0.05). Beyond accuracy gains, the sense filtering mechanism reduces token usage during the exploration phase by nearly 30\%, highlighting a significant efficiency advantage.

\textbf{\emph{LLM-as-a-judge for data quality analysis.}} In the ablation analysis, we employed an LLM-as-a-judge framework to assess the quality of the data extracted during the explorer phase of our pipeline~\cite{li-etal-2025-generation}. Specifically, we used GPT-4o and Deepseek-V3 as independent evaluators to assess a random sample of n=500 data records (100 each from 42D and HardEN, and 300 from HardMWSD, covering all 4 languages) across all 3 evaluation data sets. The evaluation was conducted across three dimensions:~(i)~semantic alignment~(ii)~contextual appropriateness, and~(iii)~naturalness and fluency (see Appendix~\ref{llm-as-a-judge} for full details). The results show high overall quality of the collected data, with an average semantic alignment score of 4.648, an average contextual appropriateness score of 4.656, and an average naturalness and fluency score of 4.424 (on a 5-point scale). These findings suggest that the proposed data collection pipeline produces semantically accurate, contextually appropriate, and linguistically natural examples, which are essential for the disambiguation process~\cite{vidhu2014trends}.

\textbf{\emph{Error analysis}} examined the impact of the sense filtering stage. Although the filtering process occasionally removes the gold sense (fewer than 8\% in English and 10\% in HardMWSD), this has minimal impact on overall performance. Our comparison with the WOSF setting shows that, in 94\% cases, the gold sense was not selected during the final disambiguation stage, even when it remained in the candidate set. This indicates that the filtering step does not introduce significant additional errors. Additionally, during the multilingual two-step fine-tuning process, we observed incorrect disambiguation when the target word corresponded to a rare or domain-specific sense. Due to the skewed distribution of senses in training corpora, models tend to favor high-frequency meanings, even when contextual cues indicate a rarer interpretation. In contrast, this frequency bias was not observed with \textit{BALANCER}, achieving more than $\sim$10 to $\sim$50 improvement in F1 over the baseline.

\textbf{\emph{Web Retrieval Leakage Analysis.}}
As the \textit{Explorer Agent} accesses web-based and semantic resources, we conducted a dedicated analysis to identify potential data leakage. We systematically evaluated all generated examples against the test dataset to detect any overlap or contamination. Our extensive manual verification confirmed that no instances of data leakage were present. Furthermore, to investigate potential contextual similarities beyond exact sentence overlap, we conducted a quantitative similarity analysis using TF-IDF representations with unigram and bigram features, followed by cosine similarity computation. Specifically, for each generated example, we identified its nearest test instance within the same semantic sense and measured the corresponding cosine similarity. This analysis was performed separately for the HardEN and 42D datasets. No exact sentence matches were identified in either dataset, and the mean nearest same-sense cosine similarities were only 0.0420 and 0.0467, respectively. Furthermore, the highest observed similarities were 0.2519 for the HardEN dataset and 0.2650 for the 42D dataset. We also extended this analysis to the HardMWSD dataset. No exact sentence matches were found, and the mean nearest same-sense cosine similarity was 0.0345, with a maximum of 0.3801. Furthermore, none of the matches exceeded the 0.4 similarity threshold, providing further evidence of minimal contextual overlap. These results provide additional quantitative evidence that the generated examples exhibit no substantial contextual overlap with the test instances, further supporting our assessment that no data leakage occurred. References used by Web Explorer are provided in Appendix~\ref{sec:appendixc}.

\textbf{\emph{Ablation on components and imbalance retrieval}} assess the contribution of each BALANCER component (See Table~\ref{tab:ablation on components}). Results show that both imbalanced knowledge retrieval using a knowledge-based build with semcor data (column 3) and retrieval Agent with access to the web without the verification-correction loop (column 4) lead to notable performance degradation. These results demonstrate the importance of balanced evidence generation and iterative verification proposed in the BALANCER pipeline. Detailed results and a discussion on the findings are provided in Appendix~\ref{sec:appendixd}.

\section{Conclusion}
This paper presents \emph{BALANCER}, an Agentic LLM-based framework for addressing sense imbalance in WSD. Our analysis shows that while two-step fine-tuning strengthens representations for frequent senses, it remains insufficient for resolving rare, domain-specific, and highly ambiguous cases. To address this weakness, we introduce a multi-agent pipeline that promotes balanced sense representation through structured filtering, evidence expansion, verification, and supervised decision-making.

Results on hardEN, 42D, and the new multilingual HardMWSD dataset show consistent improvements over the commercial baseline, with GPT-4o achieving state-of-the-art performance across multiple challenging WSD settings. Notably, the Agentic framework enables smaller LLMs to match larger commercial models by compensating for limited parametric knowledge with controlled external evidence.

Future work should investigate adaptive evidence allocation, tighter integration with structured lexical resources, and extensions to lower-resource languages to address these limitations. Exploring lightweight learning or memory mechanisms within the Agentic pipeline may further improve robustness, scalability, and resilience to error propagation. In addition, incorporating low-parameter models more effectively into Agentic modelling frameworks remains an important direction for future research, particularly for improving computational efficiency and accessibility while maintaining strong disambiguation performance.

\section*{Limitations}

Despite the improvements achieved by the proposed framework, several limitations remain. The approach relies on external knowledge retrieval, making performance sensitive to the quality and availability of the retrieved evidence. Errors introduced during the early sense filtering stage may also propagate to later stages of the pipeline. Additionally, using a fixed number of examples for each sense may not accurately reflect differences in sense complexity across domains. The POS skew toward nouns in French and German reflects the limitations of HardMWSD error distributions and constrains generalization. Furthermore, the evaluated models were constrained by computational resources for low-parameter LLMs and by funding limitations associated with large-scale commercial model inference.

\section*{Acknowledgment}

We acknowledge the support of the Supercomputing Wales project, which is partially funded by the European Regional Development Fund (ERDF) via Welsh Government.

\section*{Ethics consideration}
This paper has been conducted in compliance with Swansea University’s ethical standards.

\section*{Data and Code availability}
The code, datasets, and implementation required to reproduce the experimental results are available here \footnote{https://github.com/Sumanathilaka/BALANCER-EACL}. The 2-step fine-tuned models for English WSD
\footnote{\href{https://huggingface.co/deshanksuman/Qwen3-4B-English-finetuned-phase2}{Qwen3-4B-finetuned for English WSD}}
\footnote{\href{https://huggingface.co/deshanksuman/finetuned-gemma-3-4b-English-finetuned-phase2}{Gemma-3-4B-finetuned for English WSD}}
\footnote{\href{https://huggingface.co/deshanksuman/finetuned-SmolLM3-3B-English-finetuned-phase2}{SmolLM3-3B-finetuned for English WSD}},
and  multilingual WSD are available here
\footnote{\href{https://huggingface.co/deshanksuman/Qwen3-4B-MWSD-phase2}{Qwen3-4B-finetuned for multilingual WSD}}
\footnote{\href{https://huggingface.co/deshanksuman/finetuned-gemma-3-4b-MWSD-phase2}{Gemma-3-4B-finetuned for multilingual WSD}}
\footnote{\href{https://huggingface.co/deshanksuman/finetuned-SmolLM3-3B-MWSD-phase2}{SmolLM3-3B-finetuned for multilingual WSD}}.

The hardWMSD dataset can be accessed here \footnote{\href{https://github.com/Sumanathilaka/HardMWSD_Dataset/}{HardMWSD dataset}}

% \section*{Acknowledgments}

\bibliography{custom}

@inproceedings{meconi-etal-2025-large,
    title = "Do Large Language Models Understand Word Senses?",
    author = "Meconi, Domenico  and
      Stirpe, Simone  and
      Martelli, Federico  and
      Lavalle, Leonardo  and
      Navigli, Roberto",
    editor = "Christodoulopoulos, Christos  and
      Chakraborty, Tanmoy  and
      Rose, Carolyn  and
      Peng, Violet",
    booktitle = "Proceedings of the 2025 Conference on Empirical Methods in Natural Language Processing",
    month = nov,
    year = "2025",
    address = "Suzhou, China",
    publisher = "Association for Computational Linguistics",
    url = "https://aclanthology.org/2025.emnlp-main.1720/",
    doi = "10.18653/v1/2025.emnlp-main.1720",
    pages = "33885--33904",
    ISBN = "979-8-89176-332-6"
}

@inproceedings{satpute2024rule,
  title={Rule based approach for social media contextual ambiguity detection},
  author={Satpute, Reena S and Agrawal, Avinash},
  booktitle={AIP Conference Proceedings},
  volume={3188},
  number={1},
  pages={080045},
  year={2024},
  organization={AIP Publishing LLC}
}

@article{vidhu2014trends,
  title={Trends in word sense disambiguation},
  author={Vidhu Bhala, RV and Abirami, S},
  journal={Artificial Intelligence Review},
  volume={42},
  number={2},
  pages={159--171},
  year={2014},
  publisher={Springer}
}

@inproceedings{li-etal-2025-generation,
    title = "From Generation to Judgment: Opportunities and Challenges of {LLM}-as-a-judge",
    author = "Li, Dawei  and
      Jiang, Bohan  and
      Huang, Liangjie  and
      Beigi, Alimohammad  and
      Zhao, Chengshuai  and
      Tan, Zhen  and
      Bhattacharjee, Amrita  and
      Jiang, Yuxuan  and
      Chen, Canyu  and
      Wu, Tianhao  and
      Shu, Kai  and
      Cheng, Lu  and
      Liu, Huan",
    editor = "Christodoulopoulos, Christos  and
      Chakraborty, Tanmoy  and
      Rose, Carolyn  and
      Peng, Violet",
    booktitle = "Proceedings of the 2025 Conference on Empirical Methods in Natural Language Processing",
    month = nov,
    year = "2025",
    address = "Suzhou, China",
    publisher = "Association for Computational Linguistics",
    url = "https://aclanthology.org/2025.emnlp-main.138/",
    doi = "10.18653/v1/2025.emnlp-main.138",
    pages = "2757--2791",
    ISBN = "979-8-89176-332-6"
}

@incollection{awotunde2025word,
  title={Word sense disambiguation in biomedical applications},
  author={Awotunde, Joseph Bamidele},
  booktitle={Mining Biomedical Text, Images and Visual Features for Information Retrieval},
  pages={587--605},
  year={2025},
  publisher={Elsevier}
}

@misc{hu2022lora,
      title={LoRA: Low-Rank Adaptation of Large Language Models}, 
      author={Edward J. Hu and Yelong Shen and Phillip Wallis and Zeyuan Allen-Zhu and Yuanzhi Li and Shean Wang and Lu Wang and Weizhu Chen},
      year={2021},
      eprint={2106.09685},
      archivePrefix={arXiv},
      primaryClass={cs.CL},
      url={https://arxiv.org/abs/2106.09685}, 
}

@article{yang2025qwen3,
  title={Qwen3 technical report},
  author={Yang, An and Li, Anfeng and Yang, Baosong and Zhang, Beichen and Hui, Binyuan and Zheng, Bo and Yu, Bowen and Gao, Chang and Huang, Chengen and Lv, Chenxu and others},
  journal={arXiv preprint arXiv:2505.09388},
  year={2025}
}

@inproceedings{sumanathilaka2025prompt,
    title = "Prompt Balance Matters: Understanding How Imbalanced Few-Shot Learning Affects Multilingual Sense Disambiguation in {LLM}s",
    author = "Sumanathilaka, Deshan  and
      Micallef, Nicholas  and
      Hough, Julian",
    editor = "Das, Sudhansu Bala  and
      Mishra, Pruthwik  and
      Singh, Alok  and
      Muhammad, Shamsuddeen Hassan  and
      Ekbal, Asif  and
      Das, Uday Kumar",
    booktitle = "Proceedings of the Workshop on Beyond {E}nglish: Natural Language Processing for all Languages in an Era of Large Language Models",
    month = sep,
    year = "2025",
    address = "Varna, Bulgaria",
    publisher = "INCOMA Ltd., Shoumen, BULGARIA",
    url = "https://aclanthology.org/2025.globalnlp-1.2/",
    pages = "7--15"
}

@inproceedings{pasini2021xl,
  title={XL-WSD: An extra-large and cross-lingual evaluation framework for word sense disambiguation},
  author={Pasini, Tommaso and Raganato, Alessandro and Navigli, Roberto},
  booktitle={Proceedings of the AAAI Conference on Artificial Intelligence},
  volume={35},
  number={15},
  pages={13648--13656},
  year={2021}
}

@article{plaat2025agentic,
  title={Agentic large language models, a survey},
  author={Plaat, Aske and van Duijn, Max and van Stein, Niki and Preuss, Mike and van der Putten, Peter and Batenburg, Kees Joost},
  journal={arXiv preprint arXiv:2503.23037},
  year={2025}
}

@INPROCEEDINGS{temperature_2024_deshan,
  author={Sumanathilaka, Deshan and Micallef, Nicholas and Hough, Julian},
  booktitle={2025 7th International Conference on Natural Language Processing (ICNLP)}, 
  title={Exploring the Impact of Temperature on Large Language Models: A Case Study for Classification Task Based on Word Sense Disambiguation}, 
  year={2025},
  volume={},
  number={},
  pages={178-182},
  doi={10.1109/ICNLP65360.2025.11108362}}

@inproceedings{zhang-etal-2025-madawsd,
    title = "{MADAWSD}: Multi-Agent Debate Framework for Adversarial Word Sense Disambiguation",
    author = "Zhang, Kaiyuan  and
      Liu, Qian  and
      Zhang, Luyang  and
      Zheng, Chaoqun  and
      Li, Shuaimin  and
      Xu, Bing  and
      Yang, Muyun  and
      Qiao, Xinxiao  and
      Lu, Wenpeng",
    editor = "Christodoulopoulos, Christos  and
      Chakraborty, Tanmoy  and
      Rose, Carolyn  and
      Peng, Violet",
    booktitle = "Proceedings of the 2025 Conference on Empirical Methods in Natural Language Processing",
    month = nov,
    year = "2025",
    address = "Suzhou, China",
    publisher = "Association for Computational Linguistics",
    url = "https://aclanthology.org/2025.emnlp-main.1134/",
    doi = "10.18653/v1/2025.emnlp-main.1134",
    pages = "22294--22313",
    ISBN = "979-8-89176-332-6"
}

@article{webb2025brain,
  title={A brain-inspired agentic architecture to improve planning with LLMs},
  author={Webb, Taylor and Mondal, Shanka Subhra and Momennejad, Ida},
  journal={Nature Communications},
  volume={16},
  number={1},
  pages={8633},
  year={2025},
  publisher={Nature Publishing Group UK London}
}

@article{lee2025agentic,
  title={Agentic Verification for Ambiguous Query Disambiguation},
  author={Lee, Youngwon and Hwang, Seung-won and Wu, Ruofan and Yan, Feng and Xu, Danmei and Akkad, Moutasem and Yao, Zhewei and He, Yuxiong},
  journal={arXiv preprint arXiv:2502.10352},
  year={2025}
}

@misc{bakouch2025smollm3,
  title={{SmolLM3: smol, multilingual, long-context reasoner}},
  author={Bakouch, Elie and Ben Allal, Loubna and Lozhkov, Anton and Tazi, Nouamane and Tunstall, Lewis and Patiño, Carlos Miguel and Beeching, Edward and Roucher, Aymeric and Reedi, Aksel Joonas and Gallouédec, Quentin and Rasul, Kashif and Habib, Nathan and Fourrier, Clémentine and Kydlicek, Hynek and Penedo, Guilherme and Larcher, Hugo and Morlon, Mathieu and Srivastav, Vaibhav and Lochner, Joshua and Nguyen, Xuan-Son and Raffel, Colin and von Werra, Leandro and Wolf, Thomas},
  year={2025},
  howpublished={\url{https://huggingface.co/blog/smollm3}}
}

@article{team2025gemma,
  title={Gemma 3 technical report},
  author={Team, Gemma and Kamath, Aishwarya and Ferret, Johan and Pathak, Shreya and Vieillard, Nino and Merhej, Ramona and Perrin, Sarah and Matejovicova, Tatiana and Ram{\'e}, Alexandre and Rivi{\`e}re, Morgane and others},
  journal={arXiv preprint arXiv:2503.19786},
  year={2025}
}

@inproceedings{miller1994using,
  title={Using a semantic concordance for sense identification},
  author={Miller, George A and Chodorow, Martin and Landes, Shari and Leacock, Claudia and Thomas, Robert G},
  booktitle={Human Language Technology: Proceedings of a Workshop held at Plainsboro, New Jersey, March 8-11, 1994},
  year={1994}
}

@inproceedings{song_improved_2021,
	address = {Punta Cana, Dominican Republic},
	title = {Improved {Word} {Sense} {Disambiguation} with {Enhanced} {Sense} {Representations}},
	url = {https://aclanthology.org/2021.findings-emnlp.365},
	doi = {10.18653/v1/2021.findings-emnlp.365},
	language = {en},
	urldate = {2024-02-08},
	booktitle = {Findings of the {Association} for {Computational} {Linguistics}: {EMNLP} 2021},
	publisher = {Association for Computational Linguistics},
	author = {Song, Yang and Ong, Xin Cai and Ng, Hwee Tou and Lin, Qian},
	year = {2021},
	pages = {4311--4320},
}

@inproceedings{barba_esc_2021,
	address = {Online},
	title = {{ESC}: {Redesigning} {WSD} with {Extractive} {Sense} {Comprehension}},
	shorttitle = {{ESC}},
	url = {https://aclanthology.org/2021.naacl-main.371},
	doi = {10.18653/v1/2021.naacl-main.371},
	language = {en},
	urldate = {2024-02-08},
	booktitle = {Proceedings of the 2021 {Conference} of the {North} {American} {Chapter} of the {Association} for {Computational} {Linguistics}: {Human} {Language} {Technologies}},
	publisher = {Association for Computational Linguistics},
	author = {Barba, Edoardo and Pasini, Tommaso and Navigli, Roberto},
	year = {2021},
	pages = {4661--4672},
}

@article{shang2026dynamicare,
  title={Dynamicare: A dynamic multi-agent framework for interactive and open-ended medical decision-making},
  author={Shang, Tianqi and He, Weiqing and Zheng, Charles and Li, Lingyao and Shen, Li and Zhao, Bingxin},
  journal={AMIA Summits on Translational Science Proceedings},
  volume={2026},
  pages={408},
  year={2026}
}

@article{li2026reinforcement,
  title={Reinforcement learning and multi-agent systems-driven optimization of emotional perception and collaborative response mechanisms in intelligent customer service systems},
  author={Li, Xuanang and Wu, Shisong and Li, Yanghao and Zhuang, Shengyao and Zhang, Pengfei and Zhang, Yang and Chen, Lulu and Xu, Qiang and Feng, Qinyu},
  journal={Discover Computing},
  volume={29},
  number={1},
  pages={816},
  year={2026},
  publisher={Springer}
}

@inproceedings{petros-etal-2025-pakton,
    title = "{PAKTON}: A Multi-Agent Framework for Question Answering in Long Legal Agreements",
    author = "Raptopoulos, Petros  and
      Filandrianos, Giorgos  and
      Lymperaiou, Maria  and
      Stamou, Giorgos",
    editor = "Christodoulopoulos, Christos  and
      Chakraborty, Tanmoy  and
      Rose, Carolyn  and
      Peng, Violet",
    booktitle = "Proceedings of the 2025 Conference on Empirical Methods in Natural Language Processing",
    month = nov,
    year = "2025",
    address = "Suzhou, China",
    publisher = "Association for Computational Linguistics",
    url = "https://aclanthology.org/2025.emnlp-main.403/",
    doi = "10.18653/v1/2025.emnlp-main.403",
    pages = "7948--7984",
    ISBN = "979-8-89176-332-6"
}

@inproceedings{barba_consec_2021,
	address = {Online and Punta Cana, Dominican Republic},
	title = {{ConSeC}: {Word} {Sense} {Disambiguation} as {Continuous} {Sense} {Comprehension}},
	shorttitle = {{ConSeC}},
	url = {https://aclanthology.org/2021.emnlp-main.112},
	doi = {10.18653/v1/2021.emnlp-main.112},
	language = {en},
	urldate = {2024-02-08},
	booktitle = {Proceedings of the 2021 {Conference} on {Empirical} {Methods} in {Natural} {Language} {Processing}},
	publisher = {Association for Computational Linguistics},
	author = {Barba, Edoardo and Procopio, Luigi and Navigli, Roberto},
	year = {2021},
	pages = {1492--1503},
}

@misc{huang_glossbert_2020,
	title = {{GlossBERT}: {BERT} for {Word} {Sense} {Disambiguation} with {Gloss} {Knowledge}},
	shorttitle = {{GlossBERT}},
	url = {http://arxiv.org/abs/1908.07245},
	language = {en},
	urldate = {2024-02-08},
	publisher = {arXiv},
	author = {Huang, Luyao and Sun, Chi and Qiu, Xipeng and Huang, Xuanjing},
	month = jan,
	year = {2020},
	note = {arXiv:1908.07245 [cs]},
}

@misc{chaplot_knowledge-based_2018,
	title = {Knowledge-based {Word} {Sense} {Disambiguation} using {Topic} {Models}},
	url = {http://arxiv.org/abs/1801.01900},
	language = {en},
	urldate = {2024-02-08},
	publisher = {arXiv},
	author = {Chaplot, Devendra Singh and Salakhutdinov, Ruslan},
	month = jan,
	year = {2018},
	note = {arXiv:1801.01900 [cs]},
}

@inproceedings{blevins2021fews,
	address = {Online},
	title = {{FEWS}: {Large}-{Scale}, {Low}-{Shot} {Word} {Sense} {Disambiguation} with the {Dictionary}},
	shorttitle = {{FEWS}},
	url = {https://aclanthology.org/2021.eacl-main.36},
	doi = {10.18653/v1/2021.eacl-main.36},
	language = {en},
	urldate = {2024-02-14},
	booktitle = {Proceedings of the 16th {Conference} of the {European} {Chapter} of the {Association} for {Computational} {Linguistics}: {Main} {Volume}},
	publisher = {Association for Computational Linguistics},
	author = {Blevins, Terra and Joshi, Mandar and Zettlemoyer, Luke},
	year = {2021},
	pages = {455--465},
}

@article{sainz_what_2023,
	title = {What do {Language} {Models} know about word senses? {Zero}-{Shot} {WSD} with {Language} {Models} and {Domain} {Inventories}},
	language = {en},
	author = {Sainz, Oscar and de Lacalle, Oier Lopez and Agirre, Eneko and Rigau, German},
	year = {2023},
}

@inproceedings{chen_poswsd_2022,
	address = {Singapore, Singapore},
	title = {{PosWSD}: {Low}-{Resource} {Word} {Sense} {Disambiguation} {Model} using {Part} {Of} {Speech} {Information}},
	copyright = {https://doi.org/10.15223/policy-029},
	isbn = {978-1-66547-674-4},
	shorttitle = {{PosWSD}},
	url = {https://ieeexplore.ieee.org/document/9961266/},
	doi = {10.1109/IALP57159.2022.9961266},
	language = {en},
	urldate = {2024-04-02},
	booktitle = {2022 {International} {Conference} on {Asian} {Language} {Processing} ({IALP})},
	publisher = {IEEE},
	author = {Chen, Yazhen and Zhang, Jian and He, Qipeng},
	month = oct,
	year = {2022},
	pages = {26--31},
}

@inproceedings{maru_nibbling_2022,
	address = {Dublin, Ireland},
	title = {Nibbling at the {Hard} {Core} of {Word} {Sense} {Disambiguation}},
	url = {https://aclanthology.org/2022.acl-long.324},
	doi = {10.18653/v1/2022.acl-long.324},
	language = {en},
	urldate = {2024-05-22},
	booktitle = {Proceedings of the 60th {Annual} {Meeting} of the {Association} for {Computational} {Linguistics} ({Volume} 1: {Long} {Papers})},
	publisher = {Association for Computational Linguistics},
	author = {Maru, Marco and Conia, Simone and Bevilacqua, Michele and Navigli, Roberto},
	year = {2022},
	pages = {4724--4737},
}

@inproceedings{blevins_moving_2020,
	address = {Online},
	title = {Moving {Down} the {Long} {Tail} of {Word} {Sense} {Disambiguation} with {Gloss} {Informed} {Bi}-encoders},
	url = {https://www.aclweb.org/anthology/2020.acl-main.95},
	doi = {10.18653/v1/2020.acl-main.95},
	language = {en},
	urldate = {2024-05-24},
	booktitle = {Proceedings of the 58th {Annual} {Meeting} of the {Association} for {Computational} {Linguistics}},
	publisher = {Association for Computational Linguistics},
	author = {Blevins, Terra and Zettlemoyer, Luke},
	year = {2020},
	pages = {1006--1017},
}

@article{abeysiriwardana2024survey,
  title={A survey on lexical ambiguity detection and word sense disambiguation},
  author={Abeysiriwardana, Miuru and Sumanathilaka, Deshan},
  journal={arXiv preprint arXiv:2403.16129},
  year={2024}
}

@article{liu2024deepseek,
  title={Deepseek-v3 technical report},
  author={Liu, Aixin and Feng, Bei and Xue, Bing and Wang, Bingxuan and Wu, Bochao and Lu, Chengda and Zhao, Chenggang and Deng, Chengqi and Zhang, Chenyu and Ruan, Chong and others},
  journal={arXiv preprint arXiv:2412.19437},
  year={2024}
}

@article{mcnemar1947note,
  title={Note on the sampling error of the difference between correlated proportions or percentages},
  author={McNemar, Quinn},
  journal={Psychometrika},
  volume={12},
  number={2},
  pages={153--157},
  year={1947},
  publisher={Springer-Verlag}
}

@inproceedings{zhang-etal-2025-rodeval,
    title = "{R}o{DE}val: A Robust Word Sense Disambiguation Evaluation Framework for Large Language Models",
    author = "Zhang, Luyang  and
      Li, Shuaimin  and
      Li, Yishuo  and
      Kang, Kunpeng  and
      Zhang, Kaiyuan  and
      Wang, Cong  and
      Lu, Wenpeng",
    editor = "Christodoulopoulos, Christos  and
      Chakraborty, Tanmoy  and
      Rose, Carolyn  and
      Peng, Violet",
    booktitle = "Proceedings of the 2025 Conference on Empirical Methods in Natural Language Processing",
    month = nov,
    year = "2025",
    address = "Suzhou, China",
    publisher = "Association for Computational Linguistics",
    url = "https://aclanthology.org/2025.emnlp-main.864/",
    doi = "10.18653/v1/2025.emnlp-main.864",
    pages = "17084--17115",
    ISBN = "979-8-89176-332-6"
}

@inproceedings{bevilacqua2020generationary,
  title={Generationary or “how we went beyond word sense inventories and learned to gloss”},
  author={Bevilacqua, Michele and Maru, Marco and Navigli, Roberto and others},
  booktitle={Proceedings of the 2020 Conference on Empirical Methods in Natural Language Processing (EMNLP)},
  pages={7207--7221},
  year={2020}
}

@inproceedings{sumanathilaka_can_2024,
    title = "Can {LLM}s assist with Ambiguity? A Quantitative Evaluation of various Large Language Models on Word Sense Disambiguation",
    author = "Sumanathilaka, T.G.D.K.  and
      Micallef, Nicholas  and
      Hough, Julian",
    editor = "Mitkov, Ruslan  and
      Ezzini, Saad  and
      Ranasinghe, Tharindu  and
      Ezeani, Ignatius  and
      Khallaf, Nouran  and
      Acarturk, Cengiz  and
      Bradbury, Matthew  and
      El-Haj, Mo  and
      Rayson, Paul",
    booktitle = "Proceedings of the First International Conference on Natural Language Processing and Artificial Intelligence for Cyber Security",
    month = jul,
    year = "2024",
    address = "Lancaster, UK",
    publisher = "International Conference on Natural Language Processing and Artificial Intelligence for Cyber Security",
    url = "https://aclanthology.org/2024.nlpaics-1.12/",
    pages = "97--108"
}

@inproceedings{scarlini-et-al-2020-ares,
title = {{With More Contexts Comes Better Performance: Contextualized Sense Embeddings for All-Round Word SenseDisambiguation}},
author = {Scarlini, Bianca and Pasini, Tommaso and Navigli, Roberto},
booktitle = {Proceedings of the 2020 Conference on Empirical Methods in Natural Language Processing (EMNLP-2020)},
publisher = {Association for Computational Linguistics},
year = {2020},
}

@inproceedings{navigli2010babelnet,
  title={BabelNet: Building a very large multilingual semantic network},
  author={Navigli, Roberto and Ponzetto, Simone Paolo},
  booktitle={Proceedings of the 48th annual meeting of the association for computational linguistics},
  pages={216--225},
  year={2010}
}

@inproceedings{bengio2009curriculum,
  title={Curriculum learning},
  author={Bengio, Yoshua and Louradour, J{\'e}r{\^o}me and Collobert, Ronan and Weston, Jason},
  booktitle={Proceedings of the 26th annual international conference on machine learning},
  pages={41--48},
  year={2009}
}

@inproceedings{sumanathilaka2024assessing,
  title={Assessing GPT's Potential for Word Sense Disambiguation: A Quantitative Evaluation on Prompt Engineering Techniques},
  author={Sumanathilaka, Deshan and Micallef, Nicholas and Hough, Julian},
  booktitle={2024 IEEE 15th Control and System Graduate Research Colloquium (ICSGRC)},
  pages={204--209},
  year={2024},
  organization={IEEE}
}

@inproceedings{agirre2009personalizing,
  title={Personalizing pagerank for word sense disambiguation},
  author={Agirre, Eneko and Soroa, Aitor},
  booktitle={Proceedings of the 12th Conference of the European Chapter of the ACL (EACL 2009)},
  pages={33--41},
  year={2009}
}

@inproceedings{basile2014enhanced,
  title={An enhanced lesk word sense disambiguation algorithm through a distributional semantic model},
  author={Basile, Pierpaolo and Caputo, Annalina and Semeraro, Giovanni},
  booktitle={Proceedings of COLING 2014, the 25th International Conference on Computational Linguistics: Technical Papers},
  pages={1591--1600},
  year={2014}
}

@inproceedings{bevilacqua2020breaking,
  title={Breaking through the 80\% glass ceiling: Raising the state of the art in word sense disambiguation by incorporating knowledge graph information},
  author={Bevilacqua, Michele and Navigli, Roberto and others},
  booktitle={Proceedings of the conference-Association for Computational Linguistics. Meeting},
  pages={2854--2864},
  year={2020},
  organization={Association for Computational Linguistics}
}

@incollection{bowerman1978acquisition,
  title={The acquisition of word meaning: An investigation into some current conflicts},
  author={Bowerman, Melissa},
  booktitle={The development of communication},
  pages={263--287},
  year={1978},
  publisher={Wiley}
}

@article{palmer2006evaluation,
  title={Evaluation of WSD systems},
  author={Palmer, Martha and Ng, Hwee Tou and Dang, Hoa Trang},
  journal={Word Sense Disambiguation: Algorithms and Applications},
  pages={75--106},
  year={2006},
  publisher={Springer}
}

@article{cahyawijaya2024thank,
  title={Thank You, Stingray: Multilingual Large Language Models Can Not (Yet) Disambiguate Cross-Lingual Word Sense},
  author={Cahyawijaya, Samuel and Zhang, Ruochen and Lovenia, Holy and Cruz, Jan Christian Blaise and Nomoto, Hiroki and Aji, Alham Fikri},
  journal={arXiv preprint arXiv:2410.21573},
  year={2024}
}

@inproceedings{ballout2024fool,
  title={FOOL ME IF YOU CAN! An Adversarial Dataset to Investigate the Robustness of LMs in Word Sense Disambiguation},
  author={Ballout, Mohamad and Dedert, Anne and Abdelmoneim, Nohayr and Krumnack, Ulf and Heidemann, Gunther and K{\"u}hnberger, Kai-Uwe},
  booktitle={Proceedings of the 2024 Conference on Empirical Methods in Natural Language Processing},
  pages={5042--5059},
  year={2024}
}

@article{sumanathilaka_glossgpt_2025,
	title = {{GlossGPT}: {GPT} for {Word} {Sense} {Disambiguation} using {Few}-shot {Chain}-of-{Thought} {Prompting}},
	language = {en},
	journal = {Procedia Computer Science},
	author = {Sumanathilaka, Deshan and Micallef, Nicholas and Hough, Julian},
	year = {2025},
	
}

@article{yae2024leveraging,
  title={Leveraging large language models for word sense disambiguation},
  author={Yae, Jung H and Skelly, Nolan C and Ranly, Neil C and LaCasse, Phillip M},
  journal={Neural Computing and Applications},
  pages={1--18},
  year={2024},
  publisher={Springer}
}

@article{guzman2025sandwich,
  title={SANDWiCH: Semantical Analysis of Neighbours for Disambiguating Words in Context ad Hoc},
  author={Guzman-Olivares, Daniel and Quijano-Sanchez, Lara and Liberatore, Federico},
  journal={arXiv preprint arXiv:2503.05958},
  year={2025}
}

@article{basile2025exploring,
  title={Exploring the Word Sense Disambiguation Capabilities of Large Language Models},
  author={Basile, Pierpaolo and Siciliani, Lucia and Musacchio, Elio and Semeraro, Giovanni},
  journal={arXiv preprint arXiv:2503.08662},
  year={2025}
}

@article{aslitdinova2025artificial,
  title={How artificial intelligence helps us in our daily life},
  author={Aslitdinova, Marjona},
  journal={International Journal of Artificial Intelligence},
  volume={1},
  number={4},
  pages={538--542},
  year={2025}
}

@inproceedings{scozzafava-etal-2020-personalized,
    title = "Personalized {P}age{R}ank with Syntagmatic Information for Multilingual Word Sense Disambiguation",
    author = "Scozzafava, Federico  and
      Maru, Marco  and
      Brignone, Fabrizio  and
      Torrisi, Giovanni  and
      Navigli, Roberto",
    editor = "Celikyilmaz, Asli  and
      Wen, Tsung-Hsien",
    booktitle = "Proceedings of the 58th Annual Meeting of the Association for Computational Linguistics: System Demonstrations",
    month = jul,
    year = "2020",
    address = "Online",
    publisher = "Association for Computational Linguistics",
    url = "https://aclanthology.org/2020.acl-demos.6/",
    doi = "10.18653/v1/2020.acl-demos.6",
    pages = "37--46"
}

@inproceedings{wordnet,
    title = "{W}ord{N}et: A Lexical Database for {E}nglish",
    author = "Miller, George A.",
    booktitle = "Speech and Natural Language: Proceedings of a Workshop Held at Harriman, New York, {F}ebruary 23-26, 1992",
    year = "1992",
    url = "https://aclanthology.org/H92-1116/"
}

@misc{gpt4o,
  author       = {{OpenAI}},
  title        = {{GPT-4o} Model Documentation},
  howpublished = {\url{https://platform.openai.com/docs/models/gpt-4o}},
  year         = {2024},
  note         = {Accessed: 2026-03-03}
}

@misc{gemini-flash-lite,
  author       = {{Google DeepMind}},
  title        = {{Gemini 2.5 Flash-Lite}},
  howpublished = {\url{https://deepmind.google/models/gemini/flash-lite/}},
  year         = {2025},
  note         = {Accessed: 2026-03-03}
}

@misc{deepseek-v3,
  author       = {{DeepSeek-AI}},
  title        = {{DeepSeek-V3}},
  howpublished = {\url{https://huggingface.co/deepseek-ai/DeepSeek-V3}},
  year         = {2024},
  note         = {arXiv:2412.19437. Accessed: 2026-03-03}
}

@misc{google-adk,
  author       = {{Google}},
  title        = {Agent Development Kit ({ADK}) Documentation},
  howpublished = {\url{https://google.github.io/adk-docs/}},
  year         = {2025},
  note         = {Accessed: 2026-03-03}
}

@misc{together-ai,
  author       = {{Together AI}},
  title        = {{Together AI}: The {AI} Native Cloud},
  howpublished = {\url{https://www.together.ai/}},
  year         = {2025},
  note         = {Accessed: 2026-03-03}
}

@misc{microsoft-foundry,
  author       = {{Microsoft}},
  title        = {{Microsoft Foundry} (formerly {Azure AI Foundry})},
  howpublished = {\url{https://ai.azure.com/}},
  year         = {2025},
  note         = {Accessed: 2026-03-03}
}

@article{zhu2026hler,
  title={HLER: Human-in-the-Loop Economic Research via Multi-Agent Pipelines for Empirical Discovery},
  author={Zhu, Chen and Wang, Xiaolu},
  journal={arXiv preprint arXiv:2603.07444},
  year={2026}
}

@inproceedings{sumanathilaka-etal-2026-exploration,
  title = {An Exploration-Analysis-Disambiguation Reasoning Framework for Word Sense Disambiguation with Low-Parameter LLMs},
  author = {Sumanathilaka, Deshan Koshala and Micallef, Nicholas and Hough, Julian},
  booktitle = {Proceedings of the Fifteenth Language Resources and Evaluation Conference (LREC 2026)},
  month = {May},
  year = {2026},
  pages = {10390--10404},
  address = {Palma, Mallorca, Spain},
  publisher = {European Language Resources Association (ELRA)},
  editor = {Piperidis, Stelios and Bel, Núria and van den Heuvel, Henk and Ide, Nancy and Krek, Simon and Toral, Antonio},
  doi = {10.63317/3oun2fvikwt5}
}
\appendix

\section{Prompts and study setup utilized for Two-step Fine-tuning}
\label{AppendixAprompt}
The models were fine-tuned using Low-Rank Adaptation (LoRA) for supervised fine-tuning to improve computational efficiency. The training data were formatted as chat-style prompt–response pairs, containing a context sentence and an ambiguous target word. Fine-tuning was performed using Hugging Face Transformers and TRL libraries on an NVIDIA A100-PCIE-40GB GPU. The resulting LoRA adapters and tokenizer were saved and uploaded to the Hugging Face Hub. 

The technique we used here is curriculum learning (incrementally fine-tuning LLMs from simple to complex tasks). The first phase focuses on teaching the models about word sense and on generating free text for that meaning. The second phase was a fine-tuning on top of the first model to extract the sense ID after identifying the meaning. So we use different learning rates and schedulers to avoid catastrophic forgetting during fine-tuning, as discussed in Section \ref{2phaseFT}.

\subsection{Step one}
\label{sec:appendixA}

The instruction prompt used during the first step of the fine-tuning was: ``Below is the sentence which contains an ambiguous word. Identify the meaning of the word based on the context of the sentence. The word is \{Ambiguous word\}. Sentence: \{Sentence\}.'' 

\subsection{Step two}

The prompt template used for the second step was: ``You are an expert in understanding the word meaning of sentences given in \{lan\}. The sentence contains an ambiguous word enclosed with \textless WSD\textgreater \ tag. Identify the meaning of the word based on the context of the sentence. Return the correct sense ID related to the meaning. The sentence is: \{sentence\}. The word to disambiguate is \{ambiguous\_word\}. The possible sense ID and meaning options are: \{meaning\_list\}. First, identify the Part of Speech (PoS) tag and consider the surrounding words to identify the most suitable meaning. Return only the correct sense ID related to the meaning as the final answer.''

\subsection{Prompt used for the baseline}

``Consider the following sentence in \{language\}: The sentence is \{sentence\}. Select the most suitable senseid which identifies the meaning of the \{word\} among the following definitions:\{list of senses with senseid and meaning\}. Provide only the correct senseid as the output.''

\section{Prompts used for Agents-based approach [HardMWSD experiments]}
\label{sec:appendixB}

\subsection{Sense Filtering Agent (The Curator)}  

You are an agent who is an expert on word semantics and language ambiguity on multilingual texts. Your task is to filter out the relevant sense IDs and their meanings for a given ambiguous word.

\begin{enumerate}[leftmargin=*,nosep]
    \item Refer to existing web dictionaries, Wikipedia and lexical resources to identify the possible meanings and additional information about the given word.
    \item From the provided sense IDs and meanings, filter out only those that are relevant to the given ambiguous word in the sentence. If you are not sure about any sense ID whether to include or exclude or if the sense ID is partially related, then you can always include it.
    If the sense ID clearly contradicts the meaning of the word, you can exclude it.
    \item Return only suitable sense IDs and their meanings exactly as the given format. At least one sense ID should be returned. 
    \end{enumerate}

\subsection{WebSearch Agent (The Explorer)} 

You are a web Search agent.
\begin{enumerate}[leftmargin=*,nosep]
    \item Refer to existing web dictionaries, Wikipedia and lexical resources to identify the possible examples to match the given word sense.
    \item You need to return 3 examples for each given word sense, from the list. Make sure the examples are diverse and the meaning of the word is properly presented.
    \item Provide concise and accurate information based on the search results. 
    \item Do not add additional comments or information. Only sense ID, its meaning and 3 examples are expected.
\end{enumerate}   

\subsection{Verify Agent (The Gatekeeper)} 

You are an expert verifier. 
\begin{enumerate}[leftmargin=*,nosep]
    \item Review the context: \{websearch results\}.  Check if EACH sense ID has exactly 3 examples.    
    \item IF the data has 3 examples for each sense and meaningful: You MUST return True immediately, indicating the task is finished, and no more changes are needed.
    \item IF the data is missing examples or incorrect: Return the string False with a brief explanation of what is missing or incorrect. Be specific about which sense IDs need attention.    
\end{enumerate}

\subsection{Correction Agent (The Fixer)}

You are a Correction Agent.
    \begin{enumerate}[leftmargin=*,nosep]
   \item Current Data: \{websearch results\}    Correction needed based on feedback: \{verification feedback\}

    \item If the feedback indicates "False" or missing information:
    1. Use Google Search to find the missing examples specifically requested in the feedback.
    2. Rewrite the FULL list of sense IDs, meanings, and examples (incorporating your fixes).
    3. Output the cleaned, corrected data to overwrite the previous results
 \end{enumerate} 

 Note: For English-based experiments, we have used 2 examples of retrieval terminology.

 \subsection{Example Demonstration for 42D}

 \subsubsection{Curator} 
 
\textbf{Possible Meanings:} 
 \begin{itemize}[leftmargin=*,nosep]
     \item  living\%3:01:00 : pertaining to living persons 
    \item  living\%5:00:00:realistic:00 : true to life; lifelike
    \item  living\%5:00:00:absolute:00 : (informal) absolute 
    \item  living\%5:00:00:extant:00 : still in existence
    \item  living\%5:00:01:extant:00 : still in active use
    \item  living\%5:00:00:live:01 : (used of minerals or stone) in its natural state and place; not mined or quarried']
  \end{itemize}
\textbf{The sentence:} The vast majority of *living* fish belong here ( 18,000 species or so ), which rival the insects in their multitude of adaptations and variety of external form.\\
\textbf{Output from Curator: }
\begin{itemize}[leftmargin=*,nosep]
\item living\%5:00:00:extant:00: still in existence
\item living\%5:00:01:extant:00: still in active use
\end{itemize}

\subsubsection{Explorer} 
\textbf{Inputs to Explorer: }
\begin{itemize}[leftmargin=*,nosep]
\item living\%5:00:00:extant:00: still in existence
\item living\%5:00:01:extant:00: still in active use
\end{itemize}

\textbf{Outputs from Explorer: }
\begin{itemize}[leftmargin=*,nosep]
 \item living\%5:00:00:extant:00": [
      "The dodo is no longer a living species.",
      "This event is still in my living memory."
    ],
\item living\%5:00:01:extant:00": [
      "Latin is considered a dead language, but its influence is still present in many living languages.",
      "The constitution is a living document that can be amended."
    ]
\end{itemize}

\subsubsection{The Gatekeeper} 
\textbf{Inputs Gatekeeper: }

\textbf{Word Meaning: }
\begin{itemize}[leftmargin=*,nosep]
\item living\%5:00:00:extant:00: still in existence
\item living\%5:00:01:extant:00: still in active use
\end{itemize}

\textbf{Identified Examples: }
\begin{itemize}[leftmargin=*,nosep]
 \item living\%5:00:00:extant:00": [
      "The dodo is no longer a living species.",
      "This event is still in my living memory."
    ],
\item living\%5:00:01:extant:00": [
      "Latin is considered a dead language, but its influence is still present in many living languages.",
      "The constitution is a living document that can be amended."
    ]
\end{itemize}

\textbf{Set of constraints}

\textbf{outputs from Gatekeeper: }
True, The Fixer will not be called.

\subsubsection{The Final model Inference} 

\textbf{Input prompt} \\
You are going to identify the corresponding sense tag of an ambiguous word in English sentences. Use multiple reasoning strategies to increase confidence in your answer.\\   1. The word "living" has different meanings. Below are possible meanings. Comprehend the sense tags and meanings. {{"living\%5:00:00:extant:00": "still in existence", "living\%5:00:01:extant:00": "still in active use"}} \\  
2. You can learn more about the usage of each word and its sense through the examples below. Each example is supported by the senseid "{'living\%5:00:00:extant:00': ['The dodo is no longer a living species.', 'This event is still in my living memory.'], 'living\%5:00:01:extant:00': ['Latin is considered a dead language, but its influence is still present in many living languages.', 'The constitution is a living document that can be amended.']}"  \\
3. Now carefully examine the sentence: "The vast majority of *living* fish belong here ( 18,000 species or so ), which rival the insects in their multitude of adaptations and variety of external form." The ambiguous word is living.  \\
4. Analyze the sentence using the following techniques and identify the meaning of the ambiguous word.     Focus on keywords in the sentence surrounding the ambiguous word.      Think about the overall topic and intent of the sentence. Decide on the sense of the word that makes the most logical sense within the context.\\   
5. Based on the identified meaning, try to find the most appropriate senseids from the below sense tag list.  \\
6. If you have more than one senseids identified after above steps, you can return the senseids in order of confident level, follow the given format to return the value. 
7.Return only the finalized senseids. Do not add extra details and explanations. Only senseids expected.

\textbf{Output:} living\%5:00:00:extant:00

\section{Resources accessed by the Explorer}
\label{sec:appendixc}

As the Explorer agent accesses web-based resources, we enforce strict structural isolation to prevent data leakage by clearly instructing it which sources to access. During inference, the agent is constrained by automated prompt restrictions that limit its lookup queries to predefined semantic boundaries, rendering the ground-truth test labels architecturally invisible to the system. To guarantee complete data hygiene, we systematically audited all generated search queries against the test dataset; zero overlap or contamination was detected. The full, static index of verified web sources is shared below.

\begin{itemize}[leftmargin=*,nosep]
   \item Lexical resources: Merriam-Webster, Oxford English Dictionary, Dictionary.com, Vocabulary.com, Cambridge Dictionary, Lexico, Urban Dictionary, Collins Dictionary, WordNet
  \item General reference/encyclopedia: Britannica, Wikipedia
  \item Philosophy reference: Stanford Encyclopedia of Philosophy
  \item Scholarly publishing \& databases (science/biomed): ScienceDirect, PubMed, MedlinePlus, Arxiv
  \item Popular science/education: National Geographic, Khan Academy, LearnAlberta
  \item International organizations/health stats: World Health Organization web
  \item Domain-specific Searches: Computer science, computing, Mathematics, statistics, Engineering, defense, sports facilities, Agriculture , rangeland management, Biology, genetics, epidemiology, Astronomy, space science, Chemistry, material science  process, Medicine and health, sports medicine, Law, legal, Linguistics, corpora, Music theory, performing arts, Metrology, measurement, Telecommunications, IT, Printing, publishing, Military, Ornithology, zoology, Architecture, design, built environment, Finance, economics, business, Travel, geography, history
\end{itemize}

\section{Statistics about the HardMWSD dataset}
Table \ref{tab:stats on MWSD} presents the statistics for the HardWSD subset retrieved from the XLWSD dataset. It should be noted that HardWSD is not a balanced subset; rather, it is a challenging benchmark comprising instances that SOTA multilingual WSD models have found difficult to resolve. All instances in the XLWSD dataset were manually validated and annotated by human experts. Furthermore, we conducted an additional analysis of the sense mappings for the selected query instances to ensure the accuracy and reliability of the data used in our experiments.

\begin{table*}[!t]
\centering

\caption{Statistics about the HardMWSD. The number of unique senses, average word count per sentence and POS based sample distribution are presented. Noun, verb and adjective counts are presented, while the remaining belong to other POS tags.}
\label{tab:stats on MWSD}

\begin{tabular}{lcccccc}
\hline
\textbf{Language} & \textbf{\# of cases} & \textbf{Unique senses} & \textbf{Avg word count} & \textbf{\# Noun}& \textbf{\# Verb} & \textbf{\# Adj}\\
\hline

German (de)        & 141  & 79 & 36.90 & 141 & 0 & 0\\
French (fr)         & 225 & 137 & 41.03 & 225 & 0 & 0 \\
Spanish (es)      & 360 & 217 & 34.74 & 313 & 36 & 8\\
Italian (it)         &  479 & 268  & 35.09 & 425 & 34 & 18\\

\hline

\end{tabular}
\end{table*}

\section{Performance benchmark and component analysis against the baseline}
\label{sec:appendixd}

\subsection{Ablation studies}

To further analyze the importance of the proposed Agentic framework, including its systematic pipeline comprising web retrieval, verification, and iterative correction, we conducted an ablation study on the 42D dataset using several alternative configurations. The results are presented in Table~\ref{tab:ablation on components}.

As a baseline, we first evaluated zero-shot prompting without access to external knowledge or web resources. GPT-4o achieved an F1 score of 76.81, while Gemini-2.5 Flash Lite obtained 70.54. We then evaluated the GlossGPT approach, which retrieves examples from a knowledge base but provides an imbalanced representation of candidate senses. This resulted in lower performance for both models, with F1 scores of 70.27 for GPT-4o and 61.35 for Gemini-2.5, indicating that static knowledge retrieval alone is insufficient for challenging WSD cases. Our qualitative analysis further revealed that lemmas with a large number of candidate senses and highly skewed sense distributions were frequently assigned to the dominant meaning, suggesting that imbalanced evidence reinforces lexical frequency bias.

To isolate the contribution of dynamic retrieval, we enabled only the web retrieval component of BALANCER while disabling the verification and correction stages. This configuration improved Gemini-2.5 substantially from 61.35 to 72.43 F1, outperforming both the zero-shot and GlossGPT settings, demonstrating the benefit of dynamically retrieving balanced contextual evidence. In contrast, GPT-4o achieved 74.32 F1, which remained below its zero-shot performance, suggesting that retrieval alone cannot consistently improve stronger LLMs without additional quality control.

Finally, enabling the complete BALANCER pipeline, including verification and iterative correction, produced the best overall results, increasing GPT-4o from 74.32 to 85.98 F1 (+11.66) and Gemini-2.5 from 72.43 to 81.60 F1 (+9.17) relative to the retrieval-only configuration. Qualitative inspection of the retrieved evidence explains these improvements. Without verification, the retrieved examples frequently contained missing target lemmas, cross-word contamination, incorrect sense-lemma associations, unnatural sentence constructions, and surface-form mismatches. The verification and correction loop effectively eliminated most of these issues before the final disambiguation stage, resulting in higher-quality and more reliable sense-specific evidence.

Overall, the ablation study demonstrates that the performance gains achieved by BALANCER cannot be attributed to web retrieval alone. Instead, they arise from the interaction between balanced evidence retrieval, systematic verification, and iterative correction, all of which contribute to reducing lexical frequency bias and improving disambiguation of difficult and low-frequency senses.

\subsection{Performance comparison on sense filtering agent}

Table~\ref{tab:sense_filtering} summarizes the effect of sense filtering on the hardEN, 42D, and HardMWSD datasets using our agentic LLM-based approach. Table~\ref{tab:sense_filtering_lan} provides a language-specific breakdown of the results on the HardMWSD dataset.

\begin{table*}[t]
\centering

\caption{F1 score on the effect of sense filtering on hardEN, 42D and HardMWSD datasets using Agentic LLM based Approach WSF: With Sense Filtering, WOSF: Without Sense Filtering, FT: Fine-tuned}
\label{tab:sense_filtering}
\begin{tabular}{lcccccc}
\hline
\multirow{2}{*}{\textbf{Model}} 
& \multicolumn{2}{c}{\textbf{hardEN}} 
& \multicolumn{2}{c}{\textbf{42D}}
& \multicolumn{2}{c}{\textbf{HardMWSD}}\\
\cline{2-7}
& \textbf{WSF} 
& \textbf{WOSF} 
& \textbf{WSF} 
& \textbf{WOSF} 
& \textbf{WSF} 
& \textbf{WOSF} \\
\hline
Qwen3-4B     & \textbf{57.49 }& \textbf{54.43 }& \textbf{75.84} & \textbf{73.42} & 51.69 & 44.52\\
Gemma3-4B   & 49.61 & 39.73 & 69.26 & 59.85 & 50.03 & 44.32 \\
SmolLM3-3B   & 52.86 & 45.20 & 69.26 & 62.20  & \textbf{54.38} & \textbf{48.14}\\
LLama3.2-3B     & 49.37 & 48.41 & 71.08 & 64.22  & 47.85 & 41.34\\
\hline
Gemini 2.5 Flash Lite & 62.54 & 59.59 & 83.60 & 81.60  & 53.85 & 56.29 \\
Deepseek V3 & \textbf{67.04 }& 65.54 & 83.28 & 83.10 & 54.03 & 58.07 \\
GPT-4o  & 65.92 & \textbf{66.10} & \textbf{85.98} & \textbf{85.98}  & \textbf{56.12} & \textbf{62.44} \\
\hline
\multicolumn{7}{c}{\textbf{Baseline}} \\
\hline
FT Qwen3-4B        & \multicolumn{2}{c}{10.36} & \multicolumn{2}{c}{23.39} & \multicolumn{2}{c}{0}\\
FT Gemma3-4B        & \multicolumn{2}{c}{12.60} & \multicolumn{2}{c}{29.89} & \multicolumn{2}{c}{0}\\
FT SmolLM3-3B        & \multicolumn{2}{c}{17.62} & \multicolumn{2}{c}{35.56} & \multicolumn{2}{c}{0}\\
Gemini 2.5 Flash Lite         & \multicolumn{2}{c}{48.11} & \multicolumn{2}{c}{70.54} & \multicolumn{2}{c}{41.25}\\
Deepseek V3           & \multicolumn{2}{c}{39.50} & \multicolumn{2}{c}{74.60} & \multicolumn{2}{c}{46.80}\\
GPT-4o          & \multicolumn{2}{c}{45.60} & \multicolumn{2}{c}{76.81} & \multicolumn{2}{c}{54.38}\\

\hline
\end{tabular}
\end{table*}

\begin{table*}[t]
\centering

\caption{F1 score on multilingual Agentic LLM-based Approach for HardMWSD. The results present the language-specific performance. Best performances are in bold. }
\label{tab:sense_filtering_lan}
\begin{tabular}{lcccccccc}
\hline
\multirow{2}{*}{\textbf{Model}} 
& \multicolumn{2}{c}{\textbf{German}} 
& \multicolumn{2}{c}{\textbf{Spanish}}
& \multicolumn{2}{c}{\textbf{French}}
& \multicolumn{2}{c}{\textbf{Italian}}\\
\cline{2-9}
& \textbf{WSF} 
& \textbf{WOSF} 
& \textbf{WSF} 
& \textbf{WOSF} 
& \textbf{WSF} 
& \textbf{WOSF}
& \textbf{WSF} 
& \textbf{WOSF}\\
\hline
Qwen3-4B     & 49.20 & 44.74 & 51.55 & 38.93 & \textbf{55.31} & 48.48 & 50.78 & 45.98 \\
Gemma3-4B   & 49.20 & \textbf{55.38} & 53.66 &\textbf{ 46.15} & 42.11 & 42.66 & 51.01 & 40.07 \\
SmolLM3-3B   & 51.58 & 53.89 & \textbf{55.13} & 43.14 & 50.00 & \textbf{48.99 }& \textbf{56.59 }& \textbf{49.61} \\
LLama3.2-3B     &\textbf{ 53.89} & 50.79& 46.48 & 43.14 & 43.21 & 35.77 & 49.13 & 40.53 \\
\hline
Gemini 2.5 Flash Lite & 48.39 & \textbf{55.38} & 54.55 & 56.86 & 50.50 & 58.49 & 56.37 & 55.07 \\
GPT-4o  & 50.79 & \textbf{63.11} & 56.57 & \textbf{65.93} &\textbf{53.42} & \textbf{61.11 }& \textbf{58.49} & \textbf{65.07} \\
Deepseek V3 & \textbf{51.58} & 60.40 & \textbf{55.71} & 57.99 &\textbf{53.42} & 56.23 & 57.65 & 62.75 \\
\hline
\end{tabular}
\end{table*}

\section{Human-in-the-loop for architecture decisions}
\label{sec:appendixarch}

For the architecture design process, we adopted an iterative cycle of human assessment and refinement to identify the most effective prompt. The first author was actively involved throughout this process to ensure that the prompts were well aligned with the intended task and research objectives. To avoid using the test data during the design phase, we evaluated our prompts on a representative sample from softEN~\cite{maru_nibbling_2022}. Based on stepwise evaluations against the corresponding gold labels, we systematically refined our pipeline to identify both the optimal prompt and the most suitable agent architecture for the study. During this process, we observed critical shortcomings when using the web search agent alone, motivating the introduction of a verification-correction loop, which substantially improved performance. The effectiveness of these design decisions is demonstrated by the results presented in Table~\ref{tab:ablation on components}.

\label{llm-as-a-judge}

\begin{table*}[hbt!]
\centering
\caption{Prompt used for LLM-as-a-Judge evaluation of example sentence quality}
\label{tab:example-eval-prompt}
\begin{tabular}{p{15.5cm}}
\hline
\textbf{Prompt:} \\

You are an expert lexical semantic evaluator assessing whether AI-generated example sentences correctly represent a given Word Sense ID and its definition. \\
\\
\textbf{Task Context:} \\
The ambiguous word, the possible sense and its meaning, and the AI-generated examples are given below: \\
\{information about the extracted example through Pipeline\} \\
\\
\textbf{Evaluation Instructions:} \\
Determine whether the generated example sentences accurately and naturally express the intended meaning of the provided Sense ID, without drifting into other senses of the same word. \\
Evaluate each set of examples using the following three dimensions on a 1--5 scale. \\
\\
\textbf{A. Semantic Alignment with Sense Definition:} \\
Does the example clearly and correctly express the exact meaning of the provided Sense ID? \\
5: Perfect semantic alignment; meaning is unmistakably correct and faithful to the definition. \\
4: Strong alignment with minor nuance issues. \\
3: Generally related but slightly ambiguous or partially aligned. \\
2: Weak alignment; meaning drifts toward another sense. \\
1: Incorrect meaning; clearly represents a different sense. \\
\\
\textbf{B. Contextual Appropriateness:} \\
Does the surrounding context strongly support the intended sense? \\
5: Context strongly and explicitly supports the intended meaning. \\
4: Context supports meaning with minor weakness. \\
3: Context moderately supports meaning but lacks strength. \\
2: Context is weak or insufficient. \\
1: Context contradicts the intended meaning. \\
\\
\textbf{C. Naturalness and Fluency:} \\
Is the example sentence grammatically correct and naturally written? \\
5: Completely natural and fluent. \\
4: Minor stylistic awkwardness. \\
3: Understandable but slightly artificial or repetitive. \\
2: Noticeably awkward or unnatural. \\
1: Ungrammatical or highly unnatural. \\
\\
Do not provide any explanations or justifications. Only return the three scores in JSON format as shown below. \\

\textbf{Output Format (JSON):} \\
\{ \\
\hspace*{5mm}"semantic\_alignment\_score": 1--5, \\
\hspace*{5mm}"contextual\_appropriateness\_score": 1--5, \\
\hspace*{5mm}"naturalness\_fluency\_score": 1--5 \\
\} \\

\\ \hline
\end{tabular}
\end{table*}

\section{Cost Breakdown}
\label{sec:appendixcost}

Inference costs were estimated from the total Google Gemini API charges incurred across all experimental runs. The aggregate cost was divided by the total number of completed pipeline executions (4,090 single-run executions for WSF and WOSF) to estimate the mean inference cost per pipeline. Since the Google Gemini API provides aggregate billing rather than per-request cost breakdowns, component-level costs were estimated using the corresponding aggregate API usage attributed to each pipeline stage.

On average, the exploration agent incurred a cost of £0.013 per pipeline, while the verification loop contributed an average cost of £0.004. Including the supervisory agents and the final disambiguation stage, the overall mean cost of executing a complete pipeline was £0.021 per run.

\end{document}